\documentclass[11pt]{article}

\usepackage[utf8]{inputenc}
\usepackage[T1]{fontenc}          
\usepackage[margin=1in]{geometry}
\usepackage{microtype}

\usepackage{amsmath,amssymb,amsfonts}
\usepackage{mathrsfs}

\usepackage{graphicx}
\usepackage{xcolor}
\usepackage{textcomp}

\usepackage{booktabs}
\usepackage{tabularx}
\usepackage{array}
\usepackage{multirow}
\usepackage{adjustbox}
\usepackage{makecell}
\usepackage{ragged2e}
\usepackage[shortlabels]{enumitem}
\usepackage{caption}

\usepackage[title]{appendix}
\usepackage{listings}
\usepackage{xurl}

\usepackage[numbers,sort&compress]{natbib}

\usepackage[
  colorlinks=true,
  linkcolor=blue!45!black,
  citecolor=blue!45!black,
  urlcolor=blue!45!black,
  filecolor=blue!45!black,
  pdftitle={The Illusion of Replacement: Rethinking Specialized Machine Learning Models in the Foundation Model Era},
  pdfauthor={Kiyan Rezaee},
  pdfsubject={Survey of language-mediated versus specialized machine learning architectures},
  pdfkeywords={large language models; foundation models; structural inductive bias; architectural replacement}
]{hyperref}
\usepackage[capitalise,noabbrev]{cleveref}

\newsavebox{\wtabbox}

\newcolumntype{Y}{>{\raggedright\arraybackslash\ttfamily}X}

\newcommand{\keywords}[1]{%
  \par\vspace{0.7em}%
  \noindent\textbf{Keywords:} #1\par}

\begin{document}

\title{The Illusion of Replacement: Rethinking Specialized Machine Learning
Models in the Foundation Model Era}

\author{%
  Kiyan Rezaee \\[0.35em]
  \normalsize Department of Computer Science, University of Guilan, Rasht, Iran \\
  \normalsize \texttt{kiyanrezaee17@gmail.com}
}
\date{}

\maketitle
\thispagestyle{empty}

\begin{abstract}
\noindent
Specialized machine learning architectures encode structural assumptions (equivariance, permutation invariance, relational structure) that language-based foundation models lack by design. This review asks whether such assumptions can instead be acquired through language, drawing on a corpus of 186 papers published between 2016 and 2026 across nine modalities: tabular data, graphs, time series, vision, chemistry, code, knowledge graphs, point clouds, and protein structure. Methods are organized into eight representational regimes, ranging from language-only prompting to fully specialized architectures, and are assessed not only on predictive accuracy but on whether structural information is preserved by the representation and computed by the model. Language-mediated models prove competitive in extreme few-shot prediction, discretized symbolic tasks, and textually annotated knowledge graphs. No evidence of general replacement survives once structure itself is evaluated: where direct tests exist, apparent parity is traceable to information asymmetry, benchmark contamination, or computation performed outside the language model. Across independent research communities, missing structural inductive biases are reintroduced through graph modules, structure-aware tokens, or specialized attention, indicating that specialization is relocated rather than eliminated. The review closes by specifying a falsifiable controlled experiment that would measure the remaining gap directly. The derived data supporting the findings of this review are openly available in the repository at \url{https://github.com/kiyan-rezaee/language-vs-structure}.

\end{abstract}

\section{Introduction}
\label{sec:introduction}

Between 2024 and 2026, three research communities with no shared authorship or institutional affiliation independently made the same engineering decision. Protein-language modeling reintroduced an explicit structural channel once sequence-only representations proved insufficient: SaProt folds Foldseek-derived structure tokens\footnote{Foldseek is a fast structural-alignment tool that converts a protein's 3D backbone conformation into a short discrete alphabet, letting structural similarity be encoded and searched as if it were a sequence.} into the model's vocabulary \citep{su2024saprot}, while S-PLM aligns a sequence encoder with a separate structure encoder by contrastive learning \citep{wang2025splm}. Tabular learning kept language-model string embeddings and added an explicit graph-attention scaffold to carry relational structure \citep{kim2024carte}. Chemistry reattached graph-based tools to SMILES-based models after they proved unable to track ring systems and branching reliably \citep{bougiatiotis2026improving}. In each case a language-mediated representation was tried first, found insufficient under careful evaluation, and then supplemented with a non-linguistic structural channel. Three unrelated communities converging on the same corrective strategy is evidence of something more general about what language-based representations can and cannot carry.

The reason specialized representations existed in the first place is central to this pattern. Convolutional networks encode translation equivariance because natural images exhibit local spatial regularities \citep{cohen2016group}; permutation-invariant architectures were developed for sets because their elements have no canonical ordering \citep{lee2019set}; and message-passing networks were framed within a broader theory of relational inductive biases \citep{battaglia2018relational}. These choices encode assumptions about the hypothesis space a model should explore, and the no-free-lunch theorem explains why such assumptions are unavoidable: without assumptions about the data-generating process, no learning algorithm can outperform all others across all possible tasks \citep{wolpert1997nofreelunch}. For decades, one of the field's principal strategies was therefore to place useful structural assumptions directly into the architecture.

Large language models challenge this design paradigm. At sufficient scale, a single pretrained model can process tables, graphs, time series, and other structured inputs after they have been rendered into a language-mediated representation, often without a task-specific architecture. The important question is not whether such a model can produce a plausible prediction, but whether the structural assumptions a specialized architecture would encode explicitly are actually acquired and exploited when they are instead described, demonstrated, or implied through language. We therefore ask: \textbf{to what extent can language-mediated models acquire and exploit structural inductive biases that have traditionally been encoded explicitly in specialized machine learning architectures?}

This question cannot be answered by predictive accuracy alone. It decomposes into several questions that can have different answers. Is language a general representation for structured data, or primarily an interface to systems that perform the underlying computation? Can an inductive bias that a specialized architecture enforces by construction instead be recovered by stating or demonstrating it in a prompt? If a language-mediated model matches a specialized model's accuracy, what exactly has been replaced? And can existing evidence distinguish genuine structural generalization from benchmark contamination, encoding artifacts, or external computation? These questions motivate the framework developed throughout this review and are answered directly in \cref{sec:direct-answers}.

Read modality by modality, the empirical record appears difficult to reconcile. Positive results are real. TabLLM narrows the gap to gradient-boosted trees in the extreme few-shot regime \citep{hegselmann2023tabllm}; text-based knowledge-graph completion can outperform structural embedding baselines \citep{wang2022simkgc}; and on synthetic spatial-reasoning tasks, text-only models outperform the vision-language models given the same information as an image, so a symbolic presentation of spatial structure can outperform the pixel form \citep{wu2024is}. 

Yet controlled evaluations reveal a different picture. \citet{tan2024are} remove the language-model component from three widely cited time-series models and find that the resulting models, despite being orders of magnitude smaller and requiring up to three orders of magnitude less training and inference time, match or outperform the originals. \citet{fatemi2024talk} hold the graph, task, and model fixed while varying only the textual encoding, producing accuracy swings of 4.8--61.8 percentage points depending on the task, a variation on a property that a genuinely structure-aware representation should preserve. \citet{egressy2025setllm} start from the established fact that positional encoding and the causal attention mask make decoder-only transformers non-permutation-invariant, and prove permutation equivariance only after replacing both, meaning that a model can describe a structural property while its own computation violates it. And \citet{bordt2024elephants} show that some apparently strong few-shot tabular results cannot be separated from benchmark memorization. These findings point to different failure points in the same chain: a model may \emph{describe} a structural regularity without \emph{preserving} it, preserve information without \emph{computing} the corresponding invariant, or achieve benchmark performance without efficiently \emph{learning} the intended relationship.

This exposes a tension between \emph{generality} and \emph{structural fidelity}. A language-mediated system can be genuinely general-purpose, while failing to implement the specific invariance or compositional operation a task requires; conversely, it can match a specialized model's accuracy without matching its computational efficiency or robustness \citep{tan2024are}. The term ``replacement'' therefore conceals several non-equivalent claims. A model may functionally reproduce a specialized system's outputs without reproducing its structural computation; it may match accuracy without matching robustness; or it may orchestrate a specialized model effectively without replacing the computation that model performs. This last distinction matters increasingly, because recent work suggests language models are often most effective when they \emph{orchestrate} specialized tools and pipelines \citep{abhyankar2025llmfe,nam2025mlestar}, while specialized architectures continue to scale independently \citep{grinsztajn2025tabpfn25,qu2026tabiclv2,nxai2025tirex}.

Existing survey literature makes this distinction difficult to see, because it is organized predominantly by modality or engineering design. Surveys of language models for tabular data \citep{fang2024large,wu2025tabular}, graphs \citep{jin2023large,ren2024survey}, time series \citep{zhang2024large}, chemistry \citep{han2025from}, proteins \citep{xiao2025protein}, and point clouds \citep{thengane2025foundational} provide valuable modality-specific accounts but offer little basis for asking whether the same phenomenon recurs across domains. \citet{sun2026survey} moves toward a cross-modal perspective, organizing methods across three modalities by tokenization, architecture, pretraining, and adaptation; its taxonomy describes how systems are constructed, not what structural computations they can be shown to perform. Surveys of in-context learning \citep{dong2024survey,zhou2024mystery,mao2025survey} focus on prompting and training strategies, and geometric deep learning \citep{bronstein2021geometric} supplies the theory of specialized architectures without asking whether language can substitute for the biases they encode. To our knowledge, no existing survey systematically evaluates language-mediated methods across modalities using a common distinction between describing, preserving, computing, and learning structural properties.

This review addresses that gap across nine modalities (tabular data, graphs, time series, vision, chemistry, code, knowledge graphs, point clouds, and protein structure) through a corpus of 186 papers identified by the structured search. We separate two questions that are frequently conflated. The first concerns the \emph{representational regime}: whether structure is encoded in language alone, supplied through demonstrations or generated programs, or provided by an external specialized component; this is the eight-level taxonomy of \cref{sec:regime-ladder}. The second concerns the \emph{computational role} of that information: whether a model can \emph{describe} a structural regularity, whether its representation \emph{preserves} it, whether its forward pass \emph{computes} the corresponding invariant, and whether it can \emph{learn} it efficiently from realistic data. Systems that contain a language model but retain jointly trained or specialized components, such as Chronicle \citep{quinlan2026chronicle}, TabPFN \citep{hollmann2023tabpfn}, or PatchTST \citep{nie2023time}, are important counterfactuals to claims of language-mediated replacement, because their performance cannot be attributed to language alone.

\paragraph{Contributions.} This review makes four contributions: (i) an eight-level taxonomy of representational regimes that locates \emph{where} a system's structural computation actually happens (\cref{sec:regime-ladder}); (ii) a four-way describe/preserve/compute/learn distinction that separates \emph{what} a system does with the structural information available to it (\cref{sec:distinction}); (iii) a decomposition of ``replacement'' into eight non-equivalent senses (\cref{sec:replacement-definitions}), applied to synthesize evidence across nine modalities under one framework (\cref{sec:methods}); and (iv) the formulation of a concrete and falsifiable controlled experiment that provides a clear direction for future work (\cref{sec:open-problems}).

\paragraph{Organization.} \Cref{sec:background} describes the search methodology and positions the work against related studies. \Cref{sec:taxonomy} introduces the regime ladder, the eight senses of replacement, and the four-way distinction. \Cref{sec:methods} applies these across modalities; \cref{sec:benchmarks,sec:evaluation} ask whether current benchmarks, metrics, and theory can distinguish genuine structural generalization from contamination and format familiarity. \Cref{sec:discussion} synthesizes the cross-modal evidence, \cref{sec:open-problems} specifies the missing experiment, and \cref{sec:conclusion} concludes.

\section{Scope, Methodology, and Positioning}
\label{sec:background}

\subsection{Survey methodology}
\label{sec:methodology}

The corpus was assembled through a structured, multi-stream literature search covering 2016--2026, supplemented by a small number of earlier foundational works retained for context \citep{wolpert1997nofreelunch,wu2015shapenets}. Thirteen search streams covered tabular data, graphs, time series, vision and multimodal data, in-context learning and learning theory, universal representations, and reported limitations of language-based approaches. Each stream combined foundation-model terms with modality- and problem-specific terms; the exact queries and retrieval yields are reported in \cref{tab:search-protocol}. Backward and forward citation tracing, together with targeted searches for theoretical work and studies published in 2025--2026, complemented the database search. Retrieved records were screened using explicit inclusion and exclusion criteria. A study was retained only after its bibliographic details had been verified against a resolvable scholarly record. This procedure produced a final corpus of 186 papers.

Single-author screening introduces an important limitation: screening decisions cannot be independently reproduced by a second screener. Three instruments are therefore reported in \cref{app:audit}. First, a reproducible reference search yielded a measured sensitivity of 85.8\%. Second, the screening rule was validated against an independent reference rater, yielding $\kappa = 0.53$ and a population sensitivity of 93.8\%. Third, the 43 primary studies on which the substantive conclusions rest were assessed using SRB-6 risk-of-bias criteria. The search and screening process is additionally documented in a PRISMA-based flow diagram \citep{page2021prisma}.

\subsection{Positioning relative to existing surveys}
\label{sec:related-surveys}

A substantial survey literature already examines language and foundation models within individual modalities, including tabular data \citep{fang2024large,wu2025tabular}, graphs \citep{jin2023large,ren2024survey,liu2025graph,wang2025graphfoundation}, time series \citep{zhang2024large,jin2026large}, chemistry \citep{han2025from}, protein sequences \citep{xiao2025protein}, point clouds \citep{thengane2025foundational}, graph-based approaches to tabular data \citep{li2025graph}, and multimodal models for scientific discovery \citep{s3bench2025comprehensive}. These surveys primarily organize their respective fields by architecture, representation, adaptation strategy, or downstream task.

The present review addresses a different question: \emph{what can a language-mediated system actually be shown to compute, and what has it actually replaced?} The unit of comparison is therefore shifted from the model or modality to the computational mechanism and the evidence supporting claims about it.

Four surveys are particularly close to this perspective and therefore require explicit differentiation. \citet{sun2026survey} provides the broadest cross-modal engineering taxonomy across tabular data, time series, and graphs. It observes that language-mediated approaches can lose relational inductive bias, but does not directly test whether the underlying structural computation has been recovered. \citet{liu2025multimodal}'s \emph{TimeAsX} category identifies the specific strategy examined here, reusing a foundation model pretrained on another modality, but treats it as a design category within time-series modeling rather than as a cross-modal question about where structural computation occurs.

For graphs, \citet{yu2025graph2text} provides an especially close precedent. Its graph-to-text versus graph-to-token distinction corresponds, within graph learning, to the same representation boundary examined across nine modalities here. The distinction is generalized beyond graphs and connected to the broader question of whether structural information is merely represented for a language model or is actually computed through its language-mediated forward pass.

Finally, \citet{liang2024foundations} provides the closest structural precedent. Its framework separates representation from reasoning among six technical challenges in multimodal machine learning. That separation is retained here but made operational for the specific claim under investigation: whether language-mediated systems can replace the structural computation traditionally provided by specialized representations and architectures.

\section{A Unifying Taxonomy: Representation Regimes and the Four-Way Distinction}
\label{sec:taxonomy}

The term ``language-based'' covers systems that work in fundamentally different ways. An LLM may generate Python code that calls XGBoost, process visual tokens projected into its embedding space, or receive a structured object serialized as text. All three may be described as ``LLM-based,'' even though the structural computation occurs in different parts of the system.

We therefore distinguish two terms throughout this review. \emph{Language-based} follows the field's broader usage and is retained when reporting how prior work describes itself. \emph{Language-mediated} is our more precise term: it refers to systems in which the relevant structural computation is carried out through the language model's forward pass over language tokens. Claims made in this review's own voice use the latter.

This distinction requires two complementary taxonomies. The first asks \emph{where} the structural computation happens (\cref{sec:regime-ladder}). The second asks \emph{what the system actually does} with structural information once it has access to it (\cref{sec:distinction}). The notion of ``replacement'' lies between these two questions: to determine what a language-mediated system has replaced, we must first know where its computation occurs and then what that computation accomplishes.

\subsection{Eight representational regimes}
\label{sec:regime-ladder}

\begin{figure}[t!]
\centering
\includegraphics[width=0.85\textwidth]{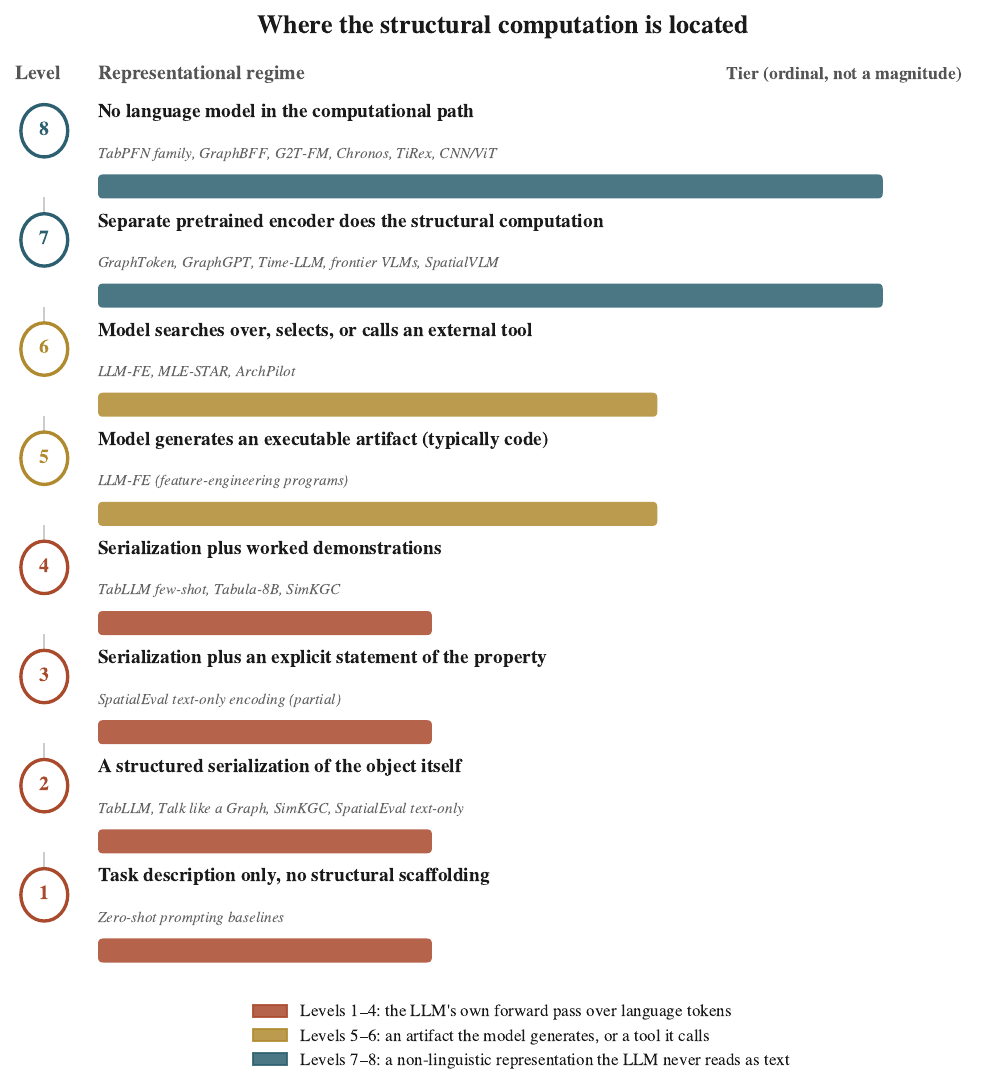}
\caption{The eight representational regimes, grouped into three tiers by where structural computation occurs. Levels 1--4 perform the computation in the language model's forward pass over natural-language tokens, with increasing linguistic scaffolding. Levels 5--6 use the language model to produce or invoke an external computational artifact. Levels 7--8 rely on non-linguistic representations or models. The boundary between Levels 6 and 7 therefore marks a fundamental shift: from language as the computational medium to a non-linguistic representation as the primary carrier of structural information.}
\label{fig:regimeladder}
\end{figure}

We classify existing approaches into eight representational regimes (\cref{fig:regimeladder}) using a single question: \emph{When a task depends on a structural property of its input, where is the computation that exploits that property performed?}

This question matters because the same input can be presented as language while the actual structural computation happens somewhere else. A graph, for example, can be serialized as text and processed by an ordinary decoder-only transformer. The fact that the input is textual does not by itself make the computation language-mediated. What matters is whether the transformer's forward pass performs the structural computation, whether the model produces an artifact that does so, whether an external tool performs it, or whether a separate encoder performs it before the language model sees the result. The eight regimes therefore form three tiers.

\paragraph{Language-mediated computation (Levels 1--4).}
At Levels 1--4, the language model's own forward pass over natural-language tokens is responsible for the structural computation. The levels differ only in how much linguistic information is provided:

\begin{itemize}
\item \textbf{L1: Task description only.} The model receives the task but no explicit structural representation.
\item \textbf{L2: Structured serialization.} The input object is converted into a textual representation.
\item \textbf{L3: Explicit structural description.} The relevant structural property is stated in natural language.
\item \textbf{L4: Worked demonstrations.} The model additionally receives examples showing how the task should be solved.
\end{itemize}

Thus, L1--L4 differ in \emph{linguistic scaffolding}, not in where the computation occurs: in all four, the language model itself is expected to perform it.

\paragraph{Language-mediated orchestration (Levels 5--6).}
At Levels 5--6, structural computation moves outside the language model's forward pass. The language model may still control the overall process, but it no longer performs the specialized computation itself.

At \textbf{L5}, the model produces an executable artifact, typically code, that carries out the computation. At \textbf{L6}, it searches over, selects, or invokes an external tool or specialized model \citep{abhyankar2025llmfe,nam2025mlestar,yuan2025archpilot}. These regimes are related to the reasoning-versus-tool-augmentation distinction used for augmented language models \citep{mialon2023augmented}, but our taxonomy makes an additional distinction that is important here: external tools are still fundamentally different from non-linguistic representations processed directly by a language model.

\paragraph{Non-linguistic representations (Levels 7--8).}
At \textbf{L7}, structural computation is performed by a separate, typically pretrained encoder before the language model processes its output. Examples include visual tokenizers, graph encoders, point-cloud encoders, and unified multimodal tokenizers \citep{tong2024eyes,tang2024graphgpt,chen2024llaga,xu2024pointllm,zhang2023metatransformer}. The language model may consume the resulting representation, but it never performs the structural transformation through language tokens.

At \textbf{L8}, the language model is absent altogether. The computation is performed by a non-linguistic model or architecture. Many L8 systems are architecturally specialized: GNNs use message passing to exploit graph structure, CNNs encode translation-related structure, and Set Transformers explicitly support permutation-invariant processing. But not every non-linguistic model is specialized in this sense.

TabPFN and related models provide an important counterexample. They contain neither a language model nor an external LLM call, but their inductive bias comes primarily from pretraining on a large distribution of synthetic tasks rather than from a modality-specific architecture. The same prior can therefore transfer, with lightweight feature engineering, to time-series forecasting \citep{hoo2025tabpfnts} and, with neighborhood-aggregated structural features, to settings in which it competes with well-tuned GNNs \citep{eremeev2025turning}. This is fundamentally different from architectural specialization, whose inductive bias is tied to a particular structural property.

For this reason, TabPFN serves in \cref{sec:methods} as the canonical strong non-linguistic baseline. The relevant comparison is therefore not simply ``language versus specialized architecture,'' but more generally ``language-mediated computation versus a strong general-purpose non-linguistic learner.''

Two boundaries in the ladder are especially important for interpreting the results. First, \textbf{L1--4 versus L6} separates language-mediated computation from language-mediated orchestration. An LLM that generates feature transformations, searches over configurations, or refines a machine-learning pipeline can effectively \emph{direct} specialized computation, but the structural computation remains in the tools it invokes. Such results demonstrate orchestration, not replacement.

Second, \textbf{L1--6 versus L7--8} separates language-mediated computation from non-linguistic representation. The presence of an LLM does not make a system language-mediated if the structural information is encoded by a separate network that the LLM never processes as language.

The regimes are not necessarily mutually exclusive within a paper. Our unit of classification is the \emph{reported result}, not the publication; the same paper can therefore contribute results at multiple levels. The tagging rule and ambiguous cases are detailed in \cref{app:boundary}.

Time-LLM \citep{jin2024timellm} illustrates why this distinction is useful. Although the method is presented as ``reprogramming'' time series through language, we classify it as L7 because its structural computation does not occur through interpretable language tokens. The taxonomy therefore makes a testable prediction: removing language should substantially disrupt a genuine L1--4 mechanism, but should leave an L7 mechanism largely intact. The ablation evidence reported in \cref{sec:timeseries-findings} is consistent with the latter prediction \citep{tan2024are}.

The same L2-versus-L7 boundary appears across modalities. Tabular methods may serialize inputs as text (L2) or use ordinary numerical in-context learning. Graph methods may use textual encodings (L2) or learned graph-encoder projections (L7), consistent with the independently proposed graph-to-text versus graph-to-token distinction of \citet{yu2025graph2text}. Vision methods similarly range from pre-discretized symbolic representations (L2--3) to learned visual tokens produced by a separate encoder (L7).

\subsection{What ``replace'' means}
\label{sec:replacement-definitions}

Identifying a system's regime still does not tell us what it has replaced. The word ``replacement'' hides several different claims. A language-mediated model can replace one property of a specialized system while failing to replace another. We therefore separate eight distinct senses of replacement (\cref{tab:replacement}). Four senses concern properties that a specialized model may provide: \textbf{functional}, \textbf{statistical}, \textbf{computational}, and \textbf{practical} replacement.

\begin{table}[t]
\centering
\small
\caption{Eight distinct senses in which a language-mediated model could be said to ``replace'' a specialized architecture (\cref{sec:taxonomy}). Treating these as one claim is the second most common overclaim identified in this review, after conflating representational levels (\cref{fig:regimeladder}).}
\label{tab:replacement}

\begin{tabularx}{\textwidth}{@{}l >{\raggedright\arraybackslash}X >{\raggedright\arraybackslash}X@{}}
\toprule
\textbf{Notion} & \textbf{Definition} & \textbf{Evidence status} \\
\midrule

Functional
& Comparable predictive performance
& Sometimes, few-shot only \\

Statistical
& Comparable sample efficiency
& Not established: sample complexity scales with parameter count and depth rather than with any structural prior \citep{yang2026tight}, and tabular accuracy degrades with dimensionality \citep{garnelo2026why} \\

Computational
& Comparable FLOPs/latency for equal accuracy
& No: up to three orders of magnitude more training/inference time for no gain \citep{tan2024are} \\

Robustness
& Comparable OOD/perturbation robustness
& No: spatial fragility \citep{saxena2026vlmrobustbench}; domain fragility in TSFMs \\

Structural
& Comparable exploitation of the relevant bias
& No: \citet{yoon2025rotor}; \citet{egressy2025setllm}; \citet{kozachinskiy2025strassen} \\

Practical
& Comparable deployment cost, latency, calibration
& No: orders of magnitude more expensive per prediction \\

Universal
& Works across broad task classes without redesign
& Partially: language's one genuine structural advantage \\

Orchestration
& Directs or composes specialized models effectively
& Yes, increasingly: a distinct Level-6 capability, and a different claim from replacement \\

\bottomrule
\end{tabularx}
\end{table}

\emph{Functional replacement} asks whether the language-mediated system achieves comparable predictive performance. \emph{Statistical replacement} asks whether it achieves that performance with comparable sample efficiency. \emph{Computational replacement} asks whether it achieves comparable accuracy with comparable computational requirements, such as FLOPs or latency. \emph{Practical replacement} asks whether it offers comparable deployment properties, including cost, latency, and calibration.

These properties are independent. Matching a specialized model's accuracy does not imply matching its sample efficiency, computational cost, or deployment characteristics. For example, \citeauthor{tan2024are}'s ablated forecasters achieve similar functional performance with orders of magnitude fewer parameters and up to three orders of magnitude less training and inference time. This is evidence against computational replacement, but it does not establish statistical replacement because sample efficiency was not evaluated under a matched-bias comparison. Two broader notions are \emph{robustness} and \emph{universality}. Robustness concerns behavior outside the training distribution, whereas universality concerns generality across task classes. Neither follows automatically from functional performance on a benchmark. 

The most important notion for this review is \emph{structural replacement}. This asks whether a language-mediated system reproduces the inductive bias that a specialized architecture would explicitly encode. Matching outputs is not enough to establish structural replacement. The reason is fundamental: an observed output does not uniquely identify the mechanism that produced it. A model can match a specialized architecture for at least four different reasons:

\begin{enumerate}
\item it genuinely performs the required structural computation;
\item it relies on pretrained knowledge rather than computing the structure \citep{bordt2024elephants};
\item its representation preserves the relevant information, while its computation does not explicitly enforce the structural property \citep{egressy2025setllm};
\item it has access to information that the specialized baseline does not \citep{wang2022simkgc}.
\end{enumerate}

Consequently, no finite collection of matched output values can, by itself, distinguish these mechanisms. Establishing structural replacement requires directly testing the relevant invariance or structural property (\cref{sec:metrics}). This concern is not merely theoretical. As \cref{sec:methods} shows, nearly every reviewed case of apparent functional or statistical replacement either fails such a structural test or has never been subjected to one.

\paragraph{Orchestration is different from replacement.}

The eighth sense, \emph{orchestration}, should not be viewed as a weaker point on the replacement scale. It describes a different relationship between the language model and specialized computation.

When a language model searches over feature transformations \citep{abhyankar2025llmfe} or refines a machine-learning pipeline \citep{nam2025mlestar}, it changes how specialized computation is selected or organized. But the specialized computation itself still occurs elsewhere. This is L6 orchestration, not replacement.

This distinction also clarifies claims that graph or tabular foundation models have already ``replaced'' specialized approaches \citep{mao2024position}. Such systems may indeed change who designs the pipeline or how models are constructed. But that is a different claim from showing that the language model itself implements the structural computation.

A related mistake is to equate the disappearance of a \emph{named} specialized architecture with the disappearance of specialization itself. In many systems, specialization has simply moved. The four examples that motivate \cref{sec:introduction} illustrate this pattern. SaProt moves structural specialization into its structure-token vocabulary; S-PLM retains it through a contrastively distilled structure encoder; CARTE incorporates it through an attention scaffold; and chemistry systems can reintroduce it through external graph tools \citep{su2024saprot,wang2025splm,kim2024carte,bougiatiotis2026improving}. In each case, the specialized component may disappear from the headline architecture, but the underlying structural machinery remains. Thus, removing a named specialized model is not evidence that specialization has been replaced. It may simply mean that \textbf{specialization has been relocated to a different part of the system}.

\subsection{The central distinction: describe, preserve, compute, learn}
\label{sec:distinction}

The regime ladder tells us \emph{where} structural computation occurs. It does not tell us \emph{what the model does} with structural information. This second question is essential because claims that a model ``understands'' a structure often combine several distinct abilities.

We separate four:

\begin{enumerate}
\item \textbf{Describe.} Can the model state or explain a structural property, such as permutation invariance or graph isomorphism? This is declarative knowledge. It is the weakest form of evidence because a model can describe a property without receiving a representation that contains the information needed to use it on a particular instance.

\item \textbf{Preserve.} After a structured object is converted into the model's representation, does that representation retain the information needed to perform the target computation? The question is not whether every detail of the original object survives, but whether the information relevant to the structural property survives.

\item \textbf{Compute.} Given a representation that contains the necessary information, does the model's forward computation actually respect the relevant structural property? A sufficient representation does not guarantee a correct computation.

\item \textbf{Learn.} Given realistic data and compute budgets, can the model acquire and generalize the relevant structural function efficiently? This concerns sample and compute efficiency, rather than whether the model can eventually learn the function.

\end{enumerate}

\emph{Describe does not imply preserve.} \citeauthor{fatemi2024talk}, for example, report swings of up to 61.8 percentage points under a fixed graph, task, and model. The model's ability to describe the relevant structure has not changed; the difference must therefore arise from what the representation preserves.

\emph{Preserve does not imply compute.} A representation may contain all the information needed for a structural computation while the model itself fails to respect the corresponding property. Standard decoder-only transformers, for example, are not permutation-invariant: positional information and causal masking make their computation order-dependent. Indeed, \citet{kazemnejad2023impact} show that the causal mask alone can allow the first layer to recover absolute positions even without explicit positional encodings. \citet{egressy2025setllm} obtain a provable guarantee only after replacing both sources of order dependence.

\emph{Compute does not require describe.} A model may implement a structural procedure without being able to explain that procedure in natural language. Mechanistic studies have identified induction mechanisms for copying, task-level representations for simple mappings, and gradient-descent-like computation in linear attention \citep{olsson2022incontext,vonoswald2023transformers,akyurek2023what}. These are examples of computation without requiring an explicit verbal description.

\emph{Learn does not require the other three.} Transformers can acquire simple function classes directly from in-context examples \citep{garg2022what}, demonstrating that learning a function need not depend on an explicit natural-language description of the underlying structure.

The distinction matters because evidence for one property is often used as evidence for another. This review treats that inference as invalid unless it is independently supported. The theory in \cref{sec:evaluation} follows the same separation: expressivity and computational limitations address \emph{compute}, whereas sample-complexity and phase-transition results address \emph{learn} \citep{kozachinskiy2025strassen,tabaghi2024universal,yang2026tight,goddard2025when}.

The two taxonomies therefore answer different questions and should not be collapsed. A Level 2 system may fail because its textual representation does not preserve the relevant information. A Level 7 system may preserve and compute that information successfully through a separate specialized encoder, while the language model contributes to neither process.

This is why both views are necessary. The regime ladder prevents us from treating the mere presence of language as evidence that language performs the computation. The four-way distinction prevents us from treating a model's ability to describe, preserve, compute, or learn one aspect of a structure as evidence that it possesses all four.

\begin{figure}[t]
\centering
\includegraphics[width=0.8\textwidth]{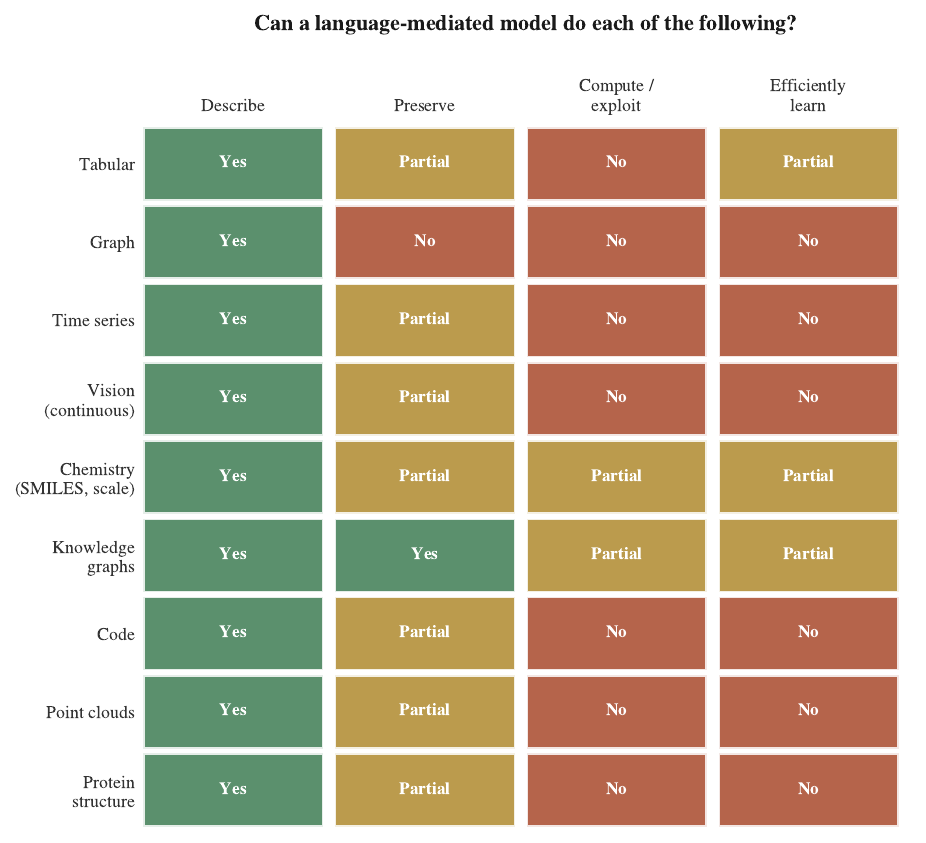}
\caption{The four-way distinction applied across the modalities reviewed in \cref{sec:methods}. Descriptive competence is common across modalities, whereas direct evidence for preservation, computation, and learning is substantially less consistent. Cells summarize the qualitative evidence synthesized in this review; no meta-analysis was performed.}
\label{fig:fourway}
\end{figure}

\Cref{fig:fourway} summarizes the resulting synthesis. The framework does not claim that language-mediated models are generally incapable of structural computation. The theoretical results reviewed in \cref{sec:evaluation} establish conditions under which they can perform such computation. The narrower empirical claim is that structural computation is \emph{task-, representation-, and architecture-dependent}. It cannot be reliably inferred from model scale, language fluency, or the ability to describe a structural property.

Accordingly, \cref{sec:methods} evaluates the four properties separately for each modality. Evidence that a model can describe a structure is not treated as evidence that it preserves that structure; preservation is not treated as evidence of computation; and computation is not treated as evidence of efficient learning.

\section{Major Methodological Families: A Modality-by-Modality Synthesis}
\label{sec:methods}

One analytical template is applied across all nine modalities in five steps: identify the specialized architecture and inductive bias traditionally used for the task; locate the language-mediated alternative on the regime ladder; weigh the strongest positive evidence against controlled, adversarial, or contamination-audited evidence that qualifies it; isolate what specialized computation remains once the language component is removed; and ask, following \cref{sec:distinction}, whether the evidence supports describing, preserving, computing, or learning the relevant structure. \Cref{tab:method-comparison} records the resulting per-system classification for twenty representative methods across the four core modalities, \cref{app:modality-detail} holds the supporting detail, and \cref{sec:methods-synthesis} draws out the mechanisms that recur.

\subsection{Tabular Data}
\label{sec:tabular-findings}

Tabular learning has long relied on tree-based ensembles and, more recently, on transformers pretrained over synthetic priors, both encoding biases suited to sparse, nonlinear feature interactions; language models are pretrained on smooth sequential structure instead. TabPFN \citep{hollmann2023tabpfn} sets a demanding non-linguistic baseline: in-context learning directly over feature--value pairs without language, outperforming boosted trees on small datasets and matching tuned AutoML systems at a reported $230\times$ end-to-end speedup. Successors extend the paradigm to datasets of up to 10{,}000 rows \citep{hollmann2025accurate,grinsztajn2025tabpfn25,qu2025tabicl,qu2026tabiclv2}, and a position paper argues the direction deserves priority in its own right \citep{vanbreugel2024why}. The comparison language-mediated methods must win is therefore against a strong general-purpose learner, not a narrow classical model.

Language-mediated approaches show their clearest advantage in the opposite regime, where labeled data are extremely scarce. TabLLM \citep{hegselmann2023tabllm} and LIFT \citep{dinh2022lift} serialize rows and prompt or few-shot the model (Levels 2 and 4), exploiting pretrained knowledge when only a handful of labels exist. The advantage is narrow and it decays: as labeled data accumulate, gradient-boosted trees overtake LLM-based prediction at substantially lower computational cost \citep{huertas2024gradient}. Stronger claims have not survived scrutiny. Tabula-8B's reported 5--15 point advantage over XGBoost and TabPFN \citep{gardner2024tabula8b} was followed by the finding that instruction-tuning the same base model with no tabular exposure reaches 92.2\% of its own mean classification accuracy, alongside train--test contamination that standard deduplication had missed \citep{gorla2026illusion}; independently, contamination has been identified in four of eight widely used tabular benchmarks, and LLMs shown to memorize popular benchmarks while performing poorly on genuinely in-context statistical tasks as dimensionality grows \citep{silvestri2026when,bordt2024elephants}. Broader evaluations narrow the positive claim further: across 142 datasets spanning IID, temporal, and grouped shift, tabular foundation models are strongest in the small-to-medium IID regime that dominates existing benchmarks, with tree-based and deep methods competitive or superior elsewhere \citep{purucker2026beyond,garnelo2026why,lafrance2026when}.

Tabular data therefore illustrates a narrow form of functional replacement, confined to a regime in which strong non-linguistic baselines do not hold the same advantage. No study in this modality tests structural replacement directly: the evidence supports descriptive competence and, under restricted conditions, functional performance, without establishing that language-mediated systems preserve or compute the inductive biases of specialized tabular learners.

\subsection{Graph-Structured Data}
\label{sec:graph-findings}

Graph neural networks encode permutation invariance by construction, from graph convolutional networks \citep{kipf2017gcn} through graph attention networks \citep{velickovic2018gat} to the more expressive Graph Isomorphism Network \citep{xu2019gin}; properly tuned classical baselines remain strong on standard node-classification benchmarks \citep{luo2024classic}. Decoder-only language models inherit no such guarantee: positional encoding and the causal mask each introduce order dependence, and \citet{egressy2025setllm} obtain a provable permutation-equivariance guarantee only after replacing both with set-specific alternatives.

The failure is empirical as well as architectural, and one pair of studies from the same group frames it unusually cleanly. Holding graph, task, and model fixed while changing only the textual encoding, \citet{fatemi2024talk} report accuracy swings of 4.8--61.8 percentage points. Their companion study takes the opposite regime, learning a parameter-efficient encoding function (GraphToken) that supplies structured information to the LLM directly and bypasses the textual-encoding choice, with gains of up to 73 points over text-only encodings on the same benchmark family \citep{perozzi2024let}. Read together, encoding sensitivity emerges as a symptom of Level 2's information bottleneck rather than a fact about language models as such: routing the structure through a Level 7 pathway removes most of it. \citet{thushalika2026detecting} likewise find reliable isomorphism detection under one node labeling but substantial degradation after relabeling the same graph, and \citet{herbst2025lost} show that fine-tuning does not simply eliminate this sensitivity.

Strong Level 7 results do not contradict this. GraphGPT and LLaGA route structural information through learned graph encoders projecting into the LLM's embedding space \citep{tang2024graphgpt,chen2024llaga}, and a unified graph foundation model trained without any language component reports comparably strong transfer \citep{liu2024one}: evidence for specialized structural encoders coupled to language models, not for natural language as the computational substrate. Comparing the two roles a language model can play, enhancing node features for a downstream GNN, or predicting directly from a serialized graph, finds the enhancer configuration consistently stronger \citep{chen2024exploring}, placing language's useful contribution at the feature level rather than at the structural computation; five further comparative evaluations agree and are collected in \cref{app:graph-detail}.

Knowledge-graph completion supplies the clearest positive result and, simultaneously, one of the clearest confounds. Text-based scoring dates at least to KG-BERT \citep{yao2019kgbert}, and SimKGC substantially extends the line, outperforming TransE, ComplEx, and RotatE \citep{wang2022simkgc}. Its inputs, however, include textual descriptions of entities and relations unavailable to the structural baselines, so the comparison introduces an information advantage and does not establish that language recovers graph topology from triples alone; consistently, \citet{yao2025exploring} find fine-tuned smaller models outperforming zero-shot frontier models on the same task.

Graphs therefore furnish the sharpest illustration of the describe/preserve distinction: models describe permutation invariance fluently while violating it in their own predictions once representation and computation are tested directly. This is one of the few modalities in which structural replacement has been tested directly, and the test fails; SimKGC's apparent functional replacement weakens once its information advantage is controlled, and the strongest graph--LLM systems retain an explicit non-linguistic pathway at Levels 7--8.

\subsection{Time-Series Forecasting}
\label{sec:timeseries-findings}

Forecasting rests on inductive biases concerning temporal order, locality, and dependence across time. Specialized architectures encode these directly, from basis-expansion networks such as N-BEATS \citep{oreshkin2020nbeats} to transformer-based forecasters, whereas a pretrained language model has no inherent reason to respect them. This modality supplies the strongest controlled evidence in the corpus against the claim that a pretrained language model automatically supplies the required temporal bias.

PromptCast established an early version of the prompt-based paradigm, reformulating forecasting as sequence-to-sequence text generation \citep{xue2023promptcast}. \citet{tan2024are} then took three widely used LLM-based forecasters,  Time-LLM \citep{jin2024timellm}, GPT4TS/OneFitsAll \citep{zhou2023one}, and CALF \citep{liu2024calf},  and systematically removed or replaced their language-model components. Across thirteen datasets and two metrics, the language-free variants outperform the originals in most comparisons while using orders of magnitude fewer parameters and up to three orders of magnitude less training and inference time. Randomly reinitializing the pretrained weights also matches or exceeds the pretrained models, and, most strikingly, shuffling the input sequence barely changes predictions despite temporal order being central to the task. The result is decisive for the three architectures tested; whether it generalizes to the broader family of LLM-adapted forecasters remains open.

Time-series foundation models reinforce the conclusion from the non-linguistic side. Chronos is framed as a ``language of time series'' but operates on quantized numerical values through a T5 architecture with no natural-language input \citep{ansari2024chronos}; TiRex uses an xLSTM architecture with no LLM component and achieves state-of-the-art results on GIFT-Eval \citep{nxai2025tirex}; and both a decoder-only patched-transformer forecaster \citep{das2024decoder} and a unified encoder trained across heterogeneous frequencies \citep{woo2024unified} report strong zero-shot performance without any language model, as do independent evaluations from epidemic forecasting to leakage audits \citep{jafari2026understanding,widener2025identifying,zhou2025why,meyer2026rethinking}. One genuinely different direction has emerged: Chronicle jointly pretrains language and time-series representations through shared attention blocks, so no pretrained language model is adapted after the fact \citep{quinlan2026chronicle}. \citeauthor{tan2024are}'s ablations do not contradict it, since it implements a different hypothesis (the two modalities learned jointly, neither attached to the other after training), though the result remains preliminary given the model's small scale and the absence of replication.

Time series therefore provides the clearest separation between describing a temporal structure and computing over it. Representing a sequence in a language-model-compatible format is trivial; the direct test of whether the language component contributes to the computation finds little evidence that it does. The verdict is correspondingly negative for computational replacement, removing the language component yields equal or better accuracy at substantially lower cost, and for structural replacement, since a perturbation that should matter to a genuinely temporal computation barely registers.

\subsection{Vision and Spatial Reasoning}
\label{sec:vision-findings}

Vision demands that three regimes be kept apart. Systems converting visual information into language before reasoning occupy Levels 2--3; vision-language models processing images through learned visual tokens connected to a language model occupy Level 7; specialized vision architectures occupy Level 8. Most current VLMs sit in the second, so their performance cannot be read as evidence that language preserves or computes visual structure.

Compositional and relational failures were documented well before current benchmarks: Winoground finds even strong models struggling to match images to captions built from identical words in different relational orders \citep{thrush2022winoground}, and a systematic follow-up traces the failure to models behaving like bags of words \citep{yuksekgonul2023when}. The pattern persists at frontier scale, with BLINK reporting 51.26\% accuracy for the best evaluated VLM across fourteen computer-vision tasks reformulated as visual question answering against 95.7\% for humans \citep{fu2024blink}, and a larger benchmark spanning depth, orientation, scale, and navigation reporting best accuracies of 54.93\% and 40.93\% against 87.57\% and 64.93\% \citep{wasi2026spatialab,rahmanzadehgervi2024vision}. Mechanistic work localizes the bottleneck rather than the symptom: predictions change relatively little when visual patches are randomly permuted \citep{lee2024intriguing}; Grid2Matrix describes a ``Digital Agnosia'' phenomenon in which the visual encoder retains substantially more information than the language output expresses \citep{zhang2026grid2matrix}; controlled variation of encoder, positional encoding, and objective across the LLaVA family leaves the deficit in place \citep{alam2026spatial}; and a minimal adapter rescaling attention-head activations recovers part of it without retraining \citep{chowdhury2026scale}. That combination is exactly the preservation-versus-computation distinction of \cref{sec:distinction}: a visual encoder may preserve spatial information the subsequent language-mediated computation fails to exploit.

Positive results nevertheless show language working well once spatial structure has already been discretized. SpatialEval finds that on synthetic maze, map, and spatial-relation tasks, text-only models outperform VLMs supplied with the same information as an image, and that multimodal models lean less on the visual channel whenever sufficient textual cues exist \citep{wu2024is}. The comparison is informative precisely because spatial content is held fixed while the channel varies, the design \cref{sec:open-problems} generalizes; \cref{app:vision-detail} discusses a further result isolating the same effect inside the visual channel itself \citep{levental2026squares}.

Vision consequently sharpens the representation/computation distinction. For raw visual inputs, structural replacement has been tested directly, through perturbations such as patch permutation, and has not been established; for already-discretized spatial structure, symbolic textual representations come closer to functional replacement, because the representation was designed to preserve what the task requires before language processing begins. Describing and reasoning over spatial relations is within reach once structure is made symbolic; preserving and exploiting continuous geometric structure through language-mediated computation remains substantially less reliable.

\subsection{Adjacent Modalities: Does the Pattern Generalize?}
\label{sec:adjacent-findings}

Five further modalities test whether the preceding pattern is specific to the four core domains. The evidence is considerably thinner, so these conclusions are directional; \cref{app:adjacent-detail} gives the extended discussion and the coverage counts behind that caveat.

\emph{Chemistry} yields the strongest positive result in the review. MoLFormer, a transformer trained on 1.1 billion molecules represented as SMILES strings with no explicit graph structure, reports outperforming supervised and self-supervised GNN baselines on several downstream tasks across ten molecular-property benchmarks \citep{ross2022large}, a genuine case of a sequence representation functionally substituting for a graph-specific architecture at sufficient scale, though the split across task types is uneven in a way the headline obscures (\cref{app:adjacent-detail}). It does not generalize within the modality: smaller SMILES-only models exhibit structural blindness to rings and branching, recovering only once graph-based information is reintroduced \citep{bougiatiotis2026improving}, and general-purpose LLMs remain insufficient for many chemistry tasks, with recent systems increasingly orchestrating specialized tools they do not replace \citep{han2025from}.

\emph{Code} offers a particularly clean test of describe-versus-compute, being already a symbolic language. CRUXEval finds frontier models reaching only 67\% and 63\% accuracy predicting the input/output behavior of short Python functions despite strong code-generation performance \citep{gu2024cruxeval}: fluent generation does not imply reliable simulation of execution. No non-linguistic specialized baseline is reported for this task, so the regime boundary for code remains untested rather than confirmed.

\emph{Point clouds} parallel the vision findings: point-cloud-augmented multimodal models outperform text-only models on some spatial-reasoning tasks yet still fail on basic binary spatial relations \citep{zhang2025point}, and PointLLM achieves strong 3D captioning by pairing a specialized point-cloud encoder with a language model \citep{xu2024pointllm}, a Level 7 configuration in which the structural representation stays in the encoder while language supplies the interface. \emph{Protein structure} reproduces the corrective move directly: SaProt \citep{su2024saprot} and S-PLM \citep{wang2025splm} reintroduce structure-derived information into sequence-based models by different mechanisms, illustrating relocated specialization rather than replacement.

Across all five, the qualification observed in the core modalities holds. Language- or sequence-based representations work well when the relevant structure is shallow, redundant with the representation, or recoverable through sufficiently large domain-specific pretraining. When the task depends critically on topology, execution semantics, or geometric relationships, the strongest systems retain or restore an explicit structural channel.

\subsection{Language as Orchestrator: The Genuine Growth Area}
\label{sec:autoML}

The strongest recent evidence for the practical value of language-mediated systems concerns orchestration, not replacement. At Level 6, LLMs generate, select, evaluate, and refine specialized computational components while the underlying structural computation stays with those components. LLM-FE generates and evaluates feature-transformation programs within an evolutionary search, consistently improving over previous automated feature-engineering methods \citep{abhyankar2025llmfe}; MLE-STAR retrieves baseline models and refines them through ablation-guided experimentation, earning medals in 64\% of evaluated Kaggle competitions \citep{nam2025mlestar}; and at industrial scale, a planner-guided multi-agent system cut feature-engineering time from three weeks to one day \citep{thakur2026towards}.

These results establish a meaningful and increasingly practical role: searching over specialized methods, composing tools, evaluating alternatives, refining pipelines. The structural computation nevertheless remains with the specialized models being invoked, so the evidence supports orchestration in the specific sense of \cref{tab:replacement} and none of the other seven. An important question follows: whether the component-ablation logic that isolated the language model's contribution in forecasting \citep{tan2024are} can determine how much of the orchestration benefit is attributable to the language model itself, and how much to the search and tool-use process it enables.

\subsection{Why these comparisons are (not) scientifically valid}
\label{sec:comparison-scientific}

Claims that one method ``outperforms'' another mean something only when the conditions of comparison are explicit, and three factors recur across every modality above. The \emph{data regime}: language-mediated methods can lead in extreme few-shot settings and lose that lead as labeled data accumulate \citep{huertas2024gradient}. The \emph{compute budget}: removing the language-model component can cut parameter count by orders of magnitude and training and inference time by up to three orders of magnitude without reducing accuracy \citep{tan2024are}, which makes accuracy-only comparisons misleading. And \emph{information availability}: SimKGC and MoLFormer both benefit from information (textual descriptions and large domain-specific pretraining corpora respectively) that may be unavailable to their specialized baselines, confounding representational choice with information access.

\Cref{tab:confound-audit} codes all three for the twelve comparisons this review relies on most, and the distribution is itself a finding: compute budget is reported or controlled in only four of twelve studies, information availability is actively \emph{violated} in four, and the comparisons treated here as strongest evidence are exactly the small minority controlling or reporting all three at once (\cref{app:confounds,app:rob}).

\subsection{Cross-modal synthesis}
\label{sec:methods-synthesis}

Taken individually, the nine modalities look like nine application areas. Taken comparatively, they converge on a much smaller set of recurring mechanisms, and those mechanisms carry the weight of this review's central argument.

The clearest convergence is a three-part pattern appearing wherever the evidence permits controlled comparison. A strong non-linguistic specialized baseline outperforms language-mediated alternatives under matched conditions: TabPFN in tabular learning, TiRex in time series, GNNs in graph learning, specialized architectures in visual reasoning. Language-mediated systems then obtain narrow advantages under particular conditions, such as extreme few-shot tabular settings or additional textual information in knowledge-graph completion. And when direct controls finally become available, either the advantage disappears under scrutiny (Tabula-8B) or the language component turns out not to be responsible for the computation (\citeauthor{tan2024are}'s ablation). A second convergence is more striking still, and \cref{sec:mechanistic-convergence} takes it up in detail: at least three modalities independently adopted the same corrective strategy of restoring an explicit non-linguistic channel once a language- or sequence-based representation failed to capture a task-relevant structural property.

The modalities diverge in a way the taxonomy predicts. Where task-relevant structure can be discretized into a symbolic representation before language processes it (SpatialEval's symbolic layouts, SMILES-based molecular representations), language models perform well, because the representation preserves much of what the task requires. Where the structure is continuous, relational, or defined by an invariance that serialization does not automatically preserve (raw visual geometry, graph topology, temporal ordering), the same strategy is substantially less reliable. \emph{Discretizability} therefore emerges as an important moderator, and a demanding one: the question is not whether a structure can be written as a sequence, but whether task-relevant structural relations survive serialization in a form the subsequent computation can exploit. Chemistry qualifies even this, since MoLFormer's success at very large scale alongside structural failures in smaller SMILES-only models suggests sufficiently large domain-specific pretraining can partially compensate for representational limitations, though whether such compensation survives matched compute, data, and information budgets is not established.

Finally, no modality provides unambiguous evidence of \emph{structural replacement} surviving a direct test, though the strength of this negative conclusion varies. Time series, graphs, and raw-pixel vision subject the claim to direct tests (component ablation, controlled serialization and permutation analysis, patch permutation), and in each case it fails. Tabular data, chemistry, code, and many benchmarks in \cref{tab:benchmarks} have undergone no equivalent test; their apparent successes may be weakened by information advantages, narrow data regimes, or contamination, but these are not direct structural tests. The evidence for this review's central claim is consequently strongest exactly where the structural property has been tested directly. \Cref{sec:discussion} returns to the harder question the modality-level evidence leaves open: whether the convergence reflects limitations of current language models or of language-mediated computation itself.

\section{Datasets and Benchmarks}
\label{sec:benchmarks}

\Cref{tab:benchmarks} summarizes the fifteen benchmarks on which this review relies most heavily, selected to illustrate what existing protocols can and cannot distinguish. Viewed through the regime ladder and the four-way distinction, a pattern emerges that is itself a finding: almost all benchmarks evaluate predictive performance, while almost none evaluates the underlying sense of replacement. TabArena, TALENT, and GIFT-Eval measure in-domain accuracy; BLINK and MMVP measure accuracy on reformatted or targeted subsets; WN18RR and FB15k-237 measure link-prediction accuracy under a comparison that is not information-matched. Only two cases depart from task-level evaluation: the component ablation of \citeauthor{tan2024are}, which is a one-off protocol rather than a standing benchmark, and the direct invariance-violation metric of \citet{yoon2025rotor} and \citet{egressy2025setllm}, discussed in \cref{sec:metrics}. Thus, across the evidence base, \emph{describe}, \emph{preserve}, \emph{compute}, and \emph{learn} are generally inferred from a single task score that was not designed to distinguish among them.

\paragraph{Benchmark bias.}
\label{sec:benchmark-bias}

Where benchmark design affects the comparison, the evidence points more strongly in one direction than the other. \emph{Regime selection} favors language-mediated methods precisely where the comparison is least informative: TabArena and TALENT are dominated by small-to-medium IID tabular datasets, the regime in which tabular foundation models perform particularly well, whereas \citet{purucker2026beyond}'s purpose-built alternative shows that this advantage weakens substantially under temporal and grouped shift. \emph{Contamination} can amplify the same advantage. Independent audits report train--test overlap in widely used tabular benchmarks \citep{gorla2026illusion,silvestri2026when} and leakage in time-series foundation-model benchmarks \citep{meyer2026rethinking}, favoring pretrained models with prior exposure to the evaluation distribution. \emph{Information asymmetry} introduces another advantage in knowledge-graph completion: WN18RR and FB15k-237 provide textual entity and relation descriptions to text-based methods that structural embedding baselines do not receive.

The reverse effect is also documented, but arises from the scoring mechanism rather than from task or information selection. A dedicated search of the evaluation-methodology literature identified systematic output-format bias across state-of-the-art LLMs on exact-match and structured-output tasks \citep{long2025biased}. Such scoring can penalize a semantically correct but format-variable answer, whereas a specialized model that produces the required format by construction avoids the same penalty. This constitutes the clearest evidence located against language-mediated methods. The two directions therefore should not be treated as offsetting biases: regime selection, contamination, and information asymmetry affect which \emph{tasks} are evaluated and which \emph{information} is available, whereas output-format bias affects how a correct answer is \emph{scored}.

Benchmark insufficiency presents a separate problem from directional bias. Of the fifteen benchmarks tabulated, five report no non-linguistic specialized-architecture baseline at all, NLGraph/GraphQA, BLINK, MMVP, MLE-Bench Lite, and CRUXEval, while a sixth remains unclear from the available sources. Results from these benchmarks therefore cannot, by themselves, establish whether a language-mediated method is competitive with the specialized computation it purports to replace.

\paragraph{What current benchmarks do not test.}

Two gaps follow directly from this evidence and motivate the open problems in \cref{sec:open-problems}. First, \emph{information preservation} and \emph{structural computation}, the second and third stages of \cref{sec:distinction}, are evaluated almost nowhere. Only two papers in the core matrix use metrics explicitly designed to detect whether a structural property is preserved but not computed; the remainder rely on accuracy measures that cannot distinguish these cases. Second, replacement in the structural or computational sense is rarely tested directly. The component ablation of \citeauthor{tan2024are} is the only genuine component-level test located, and \cref{sec:autoML} identifies the same limitation for Level-6 orchestration systems, where end-task success does not reveal how much of the result derives from the language model versus the specialized tool.

Coverage is also uneven across modalities: chemistry, protein structure, point clouds, and code each contribute only one benchmark entry, compared with several entries for the core modalities. This imbalance mirrors the thinner evidence base documented in \cref{sec:adjacent-findings} but does not resolve it. More fundamentally, no benchmark in the corpus was purpose-built to test serialization sensitivity or invariance directly. \Cref{sec:precise-gap} therefore specifies the controlled experiment motivated by this absence.

\section{Evaluation Methodology and Theoretical Foundations}
\label{sec:evaluation}

\subsection{What Current Metrics Can and Cannot Distinguish}
\label{sec:metrics}

A benchmark score is an observation of model behavior, not a measurement of the mechanism that produces it. \Cref{tab:metrics} summarizes the metrics used across the literature reviewed in \cref{sec:methods}. Read against the regime ladder and four-way distinction in \cref{sec:regime-ladder,sec:distinction}, these metrics reveal a hierarchy that benchmark practice largely leaves implicit. Accuracy, error, and task-specific scores measure \emph{output performance}: whether a prediction matches a target. Zero- and few-shot deltas measure \emph{generalization behavior}: whether performance persists with reduced adaptation. Neither measures \emph{representation fidelity} (whether the information required by the task survives its transformation into a language-mediated representation) or \emph{computational behavior} (whether the model exploits the preserved structure in the required computation). The strongest claim, \emph{mechanistic replacement}, asks whether a specialized component can be removed without loss under matched conditions, and almost none of the reviewed literature tests this directly. Two systems can therefore be predictively equivalent while differing in representational equivalence, computational equivalence, or architectural substitutability in the stricter senses of \cref{tab:replacement}. Treating a single accuracy comparison as evidence for all four is therefore a measurement-level conflation of the distinctions introduced in \cref{sec:replacement-definitions}.

Direct tests of structural behavior narrow this gap, but remain rare. A natural measure is the maximum change in model output under a structure-preserving transformation,

$$
\max_{\pi}\lvert f(x)-f(\pi(x))\rvert,
$$

which directly tests invariance. Only two studies in the reviewed literature report a related measure \citep{yoon2025rotor,egressy2025setllm}, and both use a weaker proxy: the difference between average-order and worst-case-order accuracy rather than the quantity above. The strongest available mechanistic evidence is instead a controlled ablation, removing or reinitializing a component while holding the data, task, and evaluation protocol fixed \citep{tan2024are}. Such a counterfactual comparison can establish that a component was unnecessary in a way aggregate performance cannot. Aggregate scores introduce a further limitation: averaging across datasets with different structural difficulty, as in TabArena and TALENT, can produce a favorable mean while masking failures concentrated in harder or less-represented cases \citep{purucker2026beyond}. The literature therefore measures extensively whether models are accurate, but rarely whether they perform the structural computation that could explain that accuracy.

\subsection{Why the Empirical Pattern Has a Theoretical Floor}
\label{sec:theory}

``Theoretical floor'' denotes a bounded claim rather than an impossibility result. For some structural properties, formal results establish computational or sample-complexity costs that do not disappear merely by increasing model scale within a fixed representational regime, unless the relevant structure is also supplied through architecture, representation, or training data. This is weaker than claiming that language-mediated models cannot solve such tasks in principle, but stronger than an empirical observation that current models have not yet reached sufficient scale. The distinction matters because the two interpretations make different predictions about what additional computation can achieve.

The starting point is the no-free-lunch theorem: without assumptions about the target function, no learner has a universal advantage, so reliable generalization from finite data requires an inductive bias aligned with the task structure \citep{wolpert1997nofreelunch}. This result is formal but general; it does not specify which architectures incur which costs for which structural properties. A language model also has inductive biases of its own, arising from its attention mechanism \citep{vaswani2017attention}, positional encoding, tokenization, pretraining objective, and pretraining distribution. The relevant comparison is therefore not specialized bias versus no bias, but an explicit domain-matched bias built into an architecture by construction versus a general-purpose bias whose alignment with a particular structural property is incidental and must be established empirically. Complexity-theoretic characterizations exist on both sides of this comparison \citep{lavie2024towards,carrasco2025rademacher}.

For some function classes, the match can nevertheless be favorable. \citet{kim2024transformers} prove that transformers can be minimax-optimal for classes including smooth nonparametric regression and Bayesian posterior mixtures when the pretraining distribution is matched to the target tasks. \citet{wakayama2025icl} likewise prove exponential decay of the posterior-variance component of in-context risk with the number of demonstrations under stated assumptions. These results provide formal support for the positive empirical results observed in low-data settings such as tabular prediction, but they concern specific function classes and assumptions rather than in-context learning in general. Even the mechanism underlying these successes remains contested: several studies argue that in-context regression arises because transformers implement a known algorithm during the forward pass \citep{dai2023why,xie2022explanation,bai2023transformers,zhang2024trained}, whereas \citet{hill2025transformers} report empirical evidence against that particular mechanistic account. \Cref{app:icl-theory} develops this dispute and related evidence from mathematical reasoning. The relevant point here is that theoretical guarantees for successful in-context learning do not by themselves establish why the success occurs.

For stronger compositional and relational structures, the theoretical picture is less favorable. \citet{kozachinskiy2025strassen} prove that a one-layer softmax transformer of sub-polynomial size cannot solve three-way matching, function composition, or binary-relation composition. The result holds at arbitrary numerical precision, including infinite precision, and is established using \emph{split VC dimension}, introduced by the authors for this analysis rather than classical VC dimension. Its scope is important: the barrier is precision-independent but not size-independent. The theorem applies to transformers whose size is sub-polynomial in the input length and therefore does not exclude arbitrarily wide one-layer models. Moreover, the same work constructs Strassen attention, proving that the three tasks can instead be solved with one layer and constant size. The result consequently isolates an architectural effect rather than a generic capacity limitation. It provides the clearest formal counterpart in this review to the empirical failures observed on graph-structured data (\cref{sec:graph-findings}).

Permutation invariance provides a complementary example of the same architecture-versus-scale distinction. \citet{tabaghi2024universal} prove that a permutation-invariant architecture can universally represent identifiable multiset functions with latent dimension $2DN$, where $D$ is the vector dimension and $N$ the multiset size. This substantially improves on the $O(N^D)$ bound associated with earlier sum-decomposable models \citep{takeshita2025approximation}. A standard transformer has no equivalent architectural guarantee and must instead learn permutation-invariant behavior from data if it acquires it at all. Consistent with this distinction, \citet{chiu2024deeposets} report that explicit permutation invariance can match transformer in-context performance using roughly an order of magnitude fewer parameters. These results do not show that transformers cannot learn structured functions. They show that encoding the relevant structure explicitly can yield the same performance at substantially lower parameter cost.

Two further bodies of evidence concern the learning question specifically and are developed in \cref{app:theory}. The first concerns sensitivity to superficial properties of the prompt rather than its structural content: demonstration order \citep{lu2022fantastically}, position within a long context \citep{liu2024lost}, pretraining frequency \citep{mccoy2024embers}, and even effective learning from prompts whose stated content is misleading \citep{webson2022do}. \citeauthor{yoon2025rotor} further find that prompting and formatting alone do not induce order-invariant behavior without modifying the positional mechanism. The second concerns pretraining distribution as a source of structural bias. A critical task-diversity threshold can determine whether in-context learning remains tied to the pretraining task set \citep{raventos2023pretraining,goddard2025when}, while a provable trade-off connects robustness to distribution shift with sample efficiency \citep{azizian2025how,fukuchi2026provable}. Universal-approximation results establish that transformers are expressive enough in principle to represent broad function classes \citep{li2025transformers}; however, expressivity does not imply sample or compute efficiency. No result identified here characterizes the sample complexity of learning permutation-invariant or graph-structured functions through pretraining-distribution diversity alone. This missing connection between representational expressivity, structural bias, and learning efficiency is the theoretical gap carried forward to \cref{sec:open-problems}.

\subsection{Scaling Is Real, Sub-linear, and Does Not by Itself Close the Gap}
\label{sec:scaling}

Three claims must be separated when discussing scaling: that scaling improves performance, that the improvement follows a sub-linear law with respect to a specified quantity, and that scaling closes the gap to a matched-bias specialized architecture. The literature directly supports the first, supports the second only when the scaled quantity is specified, and provides little evidence for the third. \citet{ma2025tabdpt} report a power-law exponent of approximately $0.4$ for tabular in-context learning with respect to parameter count and pretraining data size considered separately. Under $L(N)\propto N^{-\alpha}$, this corresponds to $\alpha<1$: loss decreases as scale increases, but more slowly than scale itself. Other studies report related scaling trends, but measure different quantities and therefore do not yield directly comparable exponents \citep{yao2025towards,edwards2024scaling,liu2024towards,tay2023scaling}.

None of these studies measures the quantity required by the review's central question. The relevant quantity is not the language-mediated model's loss as scale increases,

$$
L(N),
$$

but its gap relative to a matched-bias specialized architecture,

$$
\Delta(N)=L_{\text{general}}(N)-L_{\text{specialized}}(N).
$$

A decreasing $L(N)$ is compatible with $\Delta(N)$ shrinking toward zero, remaining approximately constant, or increasing, depending on how the specialized baseline scales over the same range. Indeed, \citeauthor{tay2023scaling}'s finding that the best-scaling architecture can change with model size suggests that $\Delta(N)$ need not be monotonic even when the individual loss curves are.

The evidence therefore supports a conservative conclusion: scaling reduces loss within a representational regime, and the observed reduction can follow a sub-linear scaling law, but no study identified here establishes that scaling eliminates the performance gap to a matched-bias specialized architecture. A dedicated search of the wider scaling-law literature \citep{sengupta2025how} for controlled studies that jointly scale a language-mediated model and a matched-bias specialized architecture while measuring their performance gap located no such study in either direction. The closest methodological precedent is outside language-mediated modeling. \citet{brehmer2024does} jointly scale equivariant and non-equivariant transformers on rigid-body dynamics and report the performance gap as a function of compute, finding that equivariance narrows the data-efficiency gap. This demonstrates that the proposed $\Delta(N)$ analysis is experimentally tractable. Because neither arm is language-mediated, however, the result does not answer the replacement question; it instead provides a template for testing whether scale can compensate for the absence of an explicit structural bias.

\section{Comparative Synthesis}
\label{sec:discussion}

\Cref{sec:methods} established recurring patterns across nine modalities. This section asks what those recurrences imply, answers the questions posed in \cref{sec:introduction}, and distinguishes what the evidence establishes from what it leaves unresolved.

\subsection{Is the Cross-Modality Pattern Mechanistic?}
\label{sec:mechanistic-convergence}

A similar corrective move appears independently in at least three research communities: protein-language modeling (SaProt \citep{su2024saprot} and, independently, S-PLM \citep{wang2025splm}), tabular learning (CARTE \citep{kim2024carte}), and chemistry \citep{bougiatiotis2026improving}. A direct check of author lists and institutional affiliations found no author or institutional overlap among these works. The two protein systems also arrive at the same broad correction through different mechanisms. This check cannot exclude indirect influence through shared references or broader methodological diffusion, but it supports the narrower observation that the convergence is not simply the product of one research group reproducing its own approach.

The recurrence is initially phenomenological: different communities could converge on similar remedies because they face a common constraint, because unrelated failures admit similar solutions, or because the reviewed cases disproportionately expose the same type of limitation. Several observations nevertheless favor a mechanistic interpretation. First, the corrective intervention is primarily architectural or representational rather than a consequence of scale alone: the relevant systems introduce an explicit structural channel, modify the representation, align an encoder with domain structure, or reconnect specialized computation. Increasing model size or training alone does not constitute the reported correction. Second, prompting does not appear sufficient to induce the missing structural behavior. In the order-invariance setting, \citeauthor{yoon2025rotor} find that prompting and formatting alone do not recover invariant behavior until the positional mechanism itself is modified. Third, the pattern cannot simply be attributed to weak language-based baselines. For example, MoLFormer already outperformed graph neural networks after pretraining on 1.1 billion molecules \citep{ross2022large}; the subsequent introduction of structural information therefore addresses a specific limitation within an otherwise competitive sequence-based model.

The theoretical evidence supports this interpretation, but with important scope conditions. \citeauthor{kozachinskiy2025strassen}'s split-VC-dimension result establishes a limitation for one-layer softmax attention of sub-polynomial size. Because the systems reviewed in \cref{sec:methods} are generally deeper, the theorem does not establish a general depth-independent limitation. It instead provides a formal mechanism consistent with the empirical pattern and motivates the open question in \cref{sec:rqs}, RQ1. Permutation sensitivity provides a broader architectural basis: positional encoding introduces order information, while causal masking imposes directional dependence; \citet{kazemnejad2023impact} further show that causal masking alone can recover absolute positional information. These properties arise independently of model depth. Together, the results provide two levels of theoretical support for the cross-modal pattern: a specific capacity barrier for shallow softmax attention and a more general source of order sensitivity in standard autoregressive architectures.

Methodological diffusion across neighboring fields remains possible. CARTE and SaProt, for example, draw on a broader literature on graph-based representations. The relevant claim is therefore not that independent discovery proves a common mechanism, but that diffusion alone does not explain why introducing explicit structural information repeatedly improves performance. A targeted search for cases in which an architecturally unmodified language model resolves a permutation-sensitive or compositional limitation through scale, data, or prompting alone did not identify a clear counterexample within the reviewed evidence. One important refinement emerged: \citeauthor{egressy2025setllm}'s Set-LLM restores the relevant bias internally by replacing positional encoding and causal masking with set-specific mechanisms, whereas CARTE, SaProt, and the chemistry approach introduce an external structural channel. The evidence therefore supports a more precise conclusion than ``a separate specialized component is always required'': the missing bias can be restored either externally or within the language model itself. What remains unsupported is the stronger claim that scale, additional data, or prompting alone reliably eliminates the limitation while leaving the underlying representation and computation unchanged.

\subsection{The Claim Map}
\label{sec:comparative-synthesis}

\Cref{tab:claim-map} consolidates the evidence from \cref{sec:methods,sec:evaluation} into ten recurring claims, pairing each with its strongest supporting and counter-evidence.

The strongest defensible claim concerns \emph{explicit structural information}. Several results retain both a non-linguistic representation and a task-matched architecture: PatchTST achieves strong long-horizon forecasting with a language-free patch-based architecture \citep{nie2023time}; Set Transformer explicitly encodes permutation-invariant attention \citep{lee2019set}; and GraphBFF demonstrates predictable scaling in a billion-parameter non-linguistic graph architecture \citep{becherler2026billion}. Yet TabPFN and related systems complicate a purely architectural interpretation. Their advantage over tuned boosting ensembles arises from a synthetic-prior in-context learner rather than an architecture explicitly matched to tabular structure \citep{hollmann2023tabpfn}. The same prior transfers to time-series forecasting \citep{hoo2025tabpfnts} and, with neighborhood-aggregated structural features, to settings involving graph neural networks \citep{eremeev2025turning}. The narrower and more defensible synthesis is therefore: \emph{explicit, non-linguistic structural information remains important, whether supplied by a specialized architecture or by a general-purpose in-context learner that receives appropriate structural features}. This interpretation is consistent with \cref{sec:theory}, where explicit structural bias can provide computational or statistical advantages without being tied exclusively to a particular architectural family.

A second tier contains claims that language provides a useful interface and orchestration mechanism, and that predictive competitiveness is real in specific regimes. The evidence for these claims is strong but conditional. \Cref{sec:methods} identifies settings involving extreme few-shot prediction, symbolic or discretized representations, and orchestration of specialized tools in which language-mediated systems are competitive. These results do not establish comparable performance across structural complexity, distribution shift, or matched compute budgets. A third tier concerns generalization to unseen structural tasks and statistical efficiency. Here the evidence is mixed: positive results coexist with controlled failures and with evidence of information asymmetry and contamination. These findings weaken broad interpretations without warranting the opposite claim that structural generalization or efficiency has been disproven in general. The weakest claims concern \emph{general replacement of specialized architectures}. No result identified in this review establishes such replacement while simultaneously matching representation, information availability, computation, structural invariance, and evaluation conditions.

\Cref{fig:claimbalance} summarizes this evidence balance: nine of the ten recurring claims weigh against broad substitution, while the remaining claim is itself narrower than its original framing. The resulting thesis is therefore neither that language-mediated systems cannot compete with specialized methods nor that they have generally replaced them. Rather, they can achieve behavioral competitiveness in identifiable regimes, including extreme few-shot prediction, discretized spatial representations, textually annotated knowledge graphs, and large-scale single-modality pretraining, without thereby establishing representational, computational, or architectural replacement in the stricter senses of \cref{tab:replacement}. The strongest systems increasingly reflect this distinction by using language as an interface to, or orchestrator of, specialized structural computation.

\subsection{Answers to the Central Questions}
\label{sec:direct-answers}

The confidence labels below are qualitative judgments about the evidence assembled in this review. \emph{High} indicates that multiple independent studies or a formal result directly address the claim; \emph{moderate} indicates that the evidence supports a closely related claim but does not fully establish the stated proposition; and \emph{low} indicates that the conclusion depends primarily on a single result or on the absence of counterevidence.

\textbf{Is language a general representation for structured data, or primarily an interface to systems that perform the underlying computation?} \emph{Primarily an interface} (high confidence within the structural regimes directly tested). Every domain in \cref{sec:methods} can be described, queried, or orchestrated through language, but interface universality does not imply representational sufficiency. The documented failures cluster at this boundary: language-mediated representations do not reliably preserve permutation invariance (\cref{sec:graph-findings}), temporal structure (\cref{sec:timeseries-findings}), or continuous spatial geometry (\cref{sec:vision-findings}).

\textbf{Can an inductive bias enforced by a specialized architecture be recovered by stating or demonstrating it in a prompt?} \emph{Rarely}, and not at the level of exact computational invariance in the cases directly measured (high confidence for those cases). \citeauthor{yoon2025rotor} find that prompting and formatting alone do not recover order-invariant behavior on real listwise tasks and require modification of the positional mechanism. \citet{kozachinskiy2025strassen} provide a complementary theoretical result: their one-layer limitation is a property of the computational form of attention and therefore cannot be removed merely by supplying additional instructions or demonstrations within the same constrained architecture.

\textbf{If a language-based model matches a specialized model's accuracy, what exactly has been replaced?} \emph{At most, functional behavior under the conditions of the comparison} (moderate-to-high confidence). A predictive match establishes equivalence only for the evaluated data regime, information set, and compute budget (\cref{sec:comparison-scientific}). It does not establish representational or computational equivalence. Structural replacement requires a direct invariance or information-preservation test, while mechanistic replacement requires a controlled component-level ablation. Where such tests are available in the reviewed literature, they do not establish general replacement.

\textbf{Can existing evidence distinguish genuine structural generalization from benchmark familiarity, encoding artifacts, or external computation?} \emph{Sometimes}: high confidence where a direct control exists and low confidence otherwise. Controlled ablation, contamination audits, and information-matched comparisons can separate these explanations when performed, as demonstrated by \citeauthor{tan2024are}, \citeauthor{gorla2026illusion}, and \citeauthor{fatemi2024talk}. However, \cref{sec:benchmarks} shows that such protocols remain exceptional. For most reported results, the appropriate conclusion is therefore not that structural generalization has been demonstrated or disproven, but that the evaluation does not distinguish it from the competing explanations.

\subsection{Limitations}
\label{sec:limitations}

\paragraph{The evidence base.} Coverage is uneven across modalities. The four core modalities draw on 21 distinct references for graphs, 16 for tabular data, 15 for time series, and 11 for vision, whereas the adjacent modalities contribute only three references each for chemistry and knowledge-graph completion, two each for point clouds and protein structure, and one for code. Conclusions for the adjacent modalities are therefore necessarily more tentative, and \cref{sec:adjacent-findings} presents them as directional rather than definitive. \Cref{sec:benchmarks} identifies a related limitation in evaluation: most benchmarks were not designed to test representation preservation or replacement directly. An absence of positive evidence consequently admits two readings: some claims were tested directly and failed, while others were never subjected to the relevant test. \Cref{sec:direct-answers} maintains that distinction throughout.

Evidence maturity imposes a further constraint. Of the 186 bibliography entries, 74 (40\%) are arXiv preprints without a confirmed peer-reviewed venue, and 29 (16\%) are dated 2026. Findings whose evidentiary weight depends materially on publication status are treated accordingly in the synthesis. This disclosure matters particularly for a review concerned with premature or insufficiently audited claims: part of the evidence base assembled here remains outside the peer-reviewed literature, and the risk-of-bias distribution in \cref{app:rob} should be read with that in mind. The single-author search and screening procedure is a distinct limitation, quantified rather than merely acknowledged in \cref{app:audit}.

\paragraph{The failure modes.} \Cref{tab:failure-modes} organizes the observed failures by mechanism rather than by surface symptom, which permits a more informative question: can a given failure be resolved by scale, by changing the representation, or only by restoring specialized computation? The answer differs across mechanisms. For non-permutation-invariance, positional encoding and causal ordering supply an architectural explanation, and \citeauthor{kozachinskiy2025strassen}'s result establishes a formal capacity limitation at arbitrary numerical precision for one-layer softmax attention of sub-polynomial size. Together these constitute relatively strong evidence that additional scale within the same constrained regime is not by itself a complete remedy. The remaining mechanisms require more caution. Benchmark contamination is fundamentally an evaluation problem: additional scale cannot retrospectively make a contaminated benchmark measure clean generalization, but that follows from the definition of contamination rather than from any theorem about attention. The vision-to-language bottleneck is less settled still. Discretized symbolic encodings improve performance on the tested tasks, whereas \citet{alam2026spatial} report no resolution through the encoder and objective variations they examine. That is informative controlled evidence, but it remains insufficient to establish that no alternative architectural modification could close the gap. Only some of the observed failures, therefore, currently admit a domain-agnostic mechanistic explanation.

\paragraph{The framework.} The four-way distinction of \cref{sec:distinction} is a synthesis proposed by this review. Its components correspond to distinctions already present across the theoretical and empirical literature; the contribution here is their systematic application to structured non-linguistic modalities. The framework does not, however, provide a quantitative mapping from structural complexity to the resulting performance or efficiency gap. Establishing that relationship requires controlled comparisons that vary structural complexity, representation, and scale while holding information and evaluation conditions fixed. \Cref{sec:open-problems} specifies exactly that experimental direction.

\section{Open Problems and Future Work}
\label{sec:open-problems}

\subsection{The Precise Gap}
\label{sec:precise-gap}

Existing empirical work evaluates language-mediated representations within individual modalities and inductive biases, predominantly on real-world benchmarks that confound structural competence with semantic priors, memorization, and contamination (\cref{sec:benchmark-bias}). Learning theory has independently established several relevant ingredients: sample-complexity bounds for transformers \citep{yang2026tight}, representation costs for permutation-invariant architectures \citep{tabaghi2024universal}, a split-VC-dimension ceiling for compositional tasks under one-layer attention \citep{kozachinskiy2025strassen}, and a phase transition governing structural generalization in in-context learning \citep{goddard2025when}. What is missing is the empirical link between these results: a controlled measurement of how the performance gap between a general-purpose language-mediated model and a structure-matched model changes with structural complexity and scale. The gap is therefore empirical rather than conceptual.

Let $L_{\text{general}}$ denote the loss of a language-mediated model and $L_{\text{specialized}}$ the loss of a matched architecture that encodes the relevant task structure explicitly, with training and evaluation otherwise controlled. \Cref{sec:scaling} defines

$$
\Delta(N)=L_{\text{general}}(N)-L_{\text{specialized}}(N)
$$

as the performance gap as a function of model or data scale and establishes that existing scaling studies do not measure it. The complementary cross-section is the gap as a function of structural complexity $c$ at fixed scale,

$$
\Delta(c)=L_{\text{general}}(c)-L_{\text{specialized}}(c),
$$

while holding information content and sample budget fixed and varying only the source of structural bias: architectural constraint, natural-language instruction, or in-context demonstration. The experiment should use synthetic tasks with known generating processes and the direct invariance-violation metric introduced in \cref{sec:metrics}. Together, $\Delta(N)$ and $\Delta(c)$ define a performance surface that has not been systematically measured; this review specifies the $\Delta(c)$ cross-section as the most direct first test.

The proposed gap is falsifiable. A flat or non-monotonic $\Delta(c)$ under the stated controls would weaken or refute the hypothesis that increasing structural complexity systematically exposes a growing advantage for explicit structural bias. The design is also compatible with the measurement framework already developed: the same invariance metric can quantify structural behavior while task loss measures functional performance. It provides a common axis for apparently conflicting results in \cref{sec:methods}: SimKGC and MoLFormer support strong forms of behavioral substitution in particular settings, whereas \citeauthor{tan2024are} and \citeauthor{kozachinskiy2025strassen} expose limitations under controlled structural tests. These findings occupy different points in a space that existing studies have not systematically traversed.

\subsection{Research Questions}
\label{sec:rqs}

Four questions follow, each contrasting hypotheses that make distinct empirical predictions. \Cref{app:design} gives the hypotheses, critical experiments, falsification criteria, and confounds in full, together with narrower directions derived from gaps identified in \cref{sec:methods,sec:evaluation}. The questions below state the central tests.

\textbf{RQ1: Does depth remove the compositional ceiling?} For a fixed synthetic function class, how does the performance and sample efficiency of in-context learning change relative to a matched specialized architecture as structural complexity increases? The impossibility results of \citeauthor{kozachinskiy2025strassen} apply to one-layer attention and therefore do not determine what multi-layer transformers can achieve. The competing hypotheses are directly testable: if depth removes the limitation, the gap should shrink systematically with depth; if an architectural ceiling persists, the gap should remain under controlled parameter and compute budgets.

\textbf{RQ2: Does stating an invariance in language reduce its violation?} When a structural invariance is explicitly stated, does the model's output become measurably more invariant under the direct metric of \cref{sec:metrics}? Existing studies show that natural-language instructions can alter learned behavior while strong pretraining priors remain difficult to override \citep{si2023measuring}, and that the same concept can produce different behavior depending on whether it is conveyed through image or text \citep{allen2025incontext}. Neither directly measures structural invariance across a complexity sweep against a matched specialized baseline. RQ2 therefore tests a specific unresolved question: whether language can communicate an inductive bias at the level of the computation itself rather than merely shift task behavior.

\textbf{RQ3: Do demonstrations close more of the gap than instructions, and can their effect be separated from memorization?} \citeauthor{wakayama2025icl}'s exponential-decay result predicts rapidly decreasing risk when the target structure lies within the support of the pretraining task mixture, whereas \citeauthor{goddard2025when}'s phase transition predicts a qualitative degradation when it falls outside that support, with a reported threshold near $120^\circ$ of task-angle diversity (approximately $135^\circ$ under label noise). The unresolved question is whether the same phenomenon appears when diversity is indexed by structural complexity rather than task angle. A controlled experiment can therefore vary demonstration diversity while holding the underlying generating rule and test distribution fixed, separating structural learning from memorization.

\textbf{RQ4: What predicts the gap in advance?} Which measurable properties of a task, such as symmetry-group order, minimum description length, interaction structure, or discretizability, predict whether $\Delta(c)$ remains small or grows rapidly? The discrete-versus-continuous distinction in \cref{sec:vision-findings}, where symbolic spatial grids outperform pixel representations on matched tasks while continuous geometry remains difficult, suggests that discretizability may be an important moderator. The $2DN$ representation bound of \citeauthor{tabaghi2024universal} provides a principled complexity variable against which such predictors can be tested.

\section{Conclusion}
\label{sec:conclusion}

Across nine modalities and four methodological lenses, one conclusion emerges: \emph{behavioral competitiveness is not replacement}. Language-mediated systems can match specialized architectures in predictive performance, but when representation, computation, or architecture is tested directly, such parity rarely extends beyond the narrower senses identified in \cref{tab:replacement}. Language can substitute for explicit inductive bias in identifiable regimes, including extreme few-shot prediction, discretizable spatial layouts, textually annotated knowledge graphs, and settings in which sufficiently large pretraining compensates for limited task-specific structure. Outside these regimes, apparent success is often entangled with contamination, format familiarity, or information supplied through channels that are not purely linguistic. The methodological implication is consequential: accuracy, statistical efficiency, computational cost, and structural equivalence are distinct claims, and benchmark performance alone establishes only the first.

A recurring cross-modal pattern further emerges. When language-only representations prove insufficient, successful systems tend to reintroduce non-linguistic structure through specialized tokens, external tools, or architectural scaffolds. What appears to be the disappearance of specialization is therefore often its relocation. This pattern recurs across independent research communities, although the available evidence does not establish it as a universal law. Language also provides an effective interface for directing and composing specialized components, but such integration constitutes a different form of substitution from architectural replacement. Scaling presents the same distinction: the relevant question is not whether $L(N)$ decreases with scale, but whether the performance gap to a matched-bias architecture, $\Delta(N)$, decreases and eventually closes (\cref{sec:scaling}).

The central question is consequently more precise than whether language can replace specialized machine learning. Existing theoretical results provide important constraints in specific settings, including the compositional-task split-VC-dimension ceiling for one-layer attention established by \citet{kozachinskiy2025strassen} and the pretraining-diversity phase transition reported by \citet{goddard2025when}, but neither predicts the performance gap as a function of scale and structural complexity (\cref{sec:precise-gap,sec:scaling}). The field therefore needs a quantitative account that predicts, in advance, where language-mediated systems cease to substitute effectively for specialized structure.

\begin{appendices}


\section*{Appendix}

\begin{figure}[t]
\centering
\includegraphics[width=\textwidth]{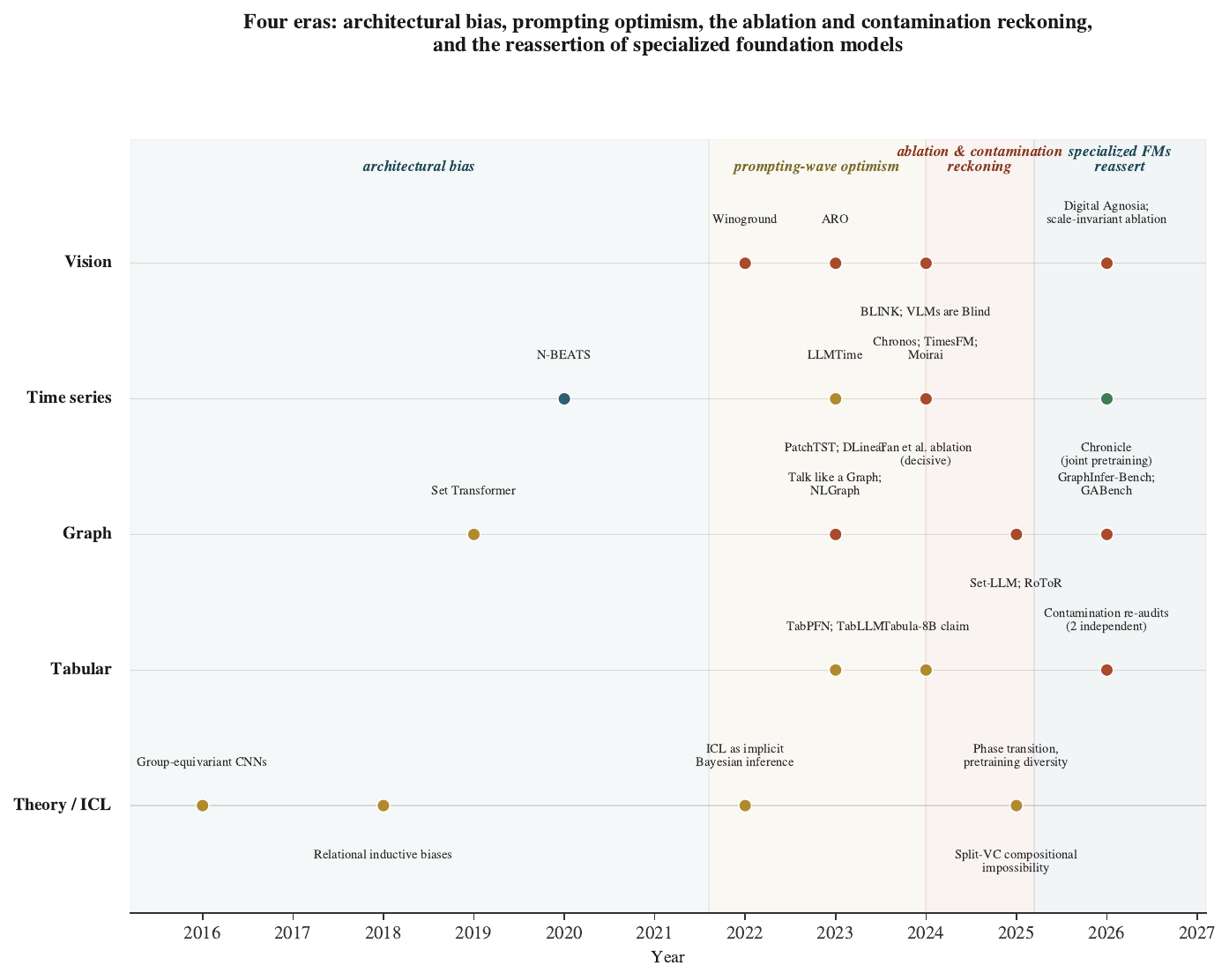}
\caption{Historical evolution of research on language-mediated learning of structured data, 2016--2026: from explicitly encoded structural inductive biases, through language-mediated representations and prompting, to controlled ablations, contamination analyses, theoretical investigations, and renewed interest in specialized and jointly pretrained foundation models. Era boundaries are approximate and overlap.}
\label{fig:timeline}
\end{figure}

\begin{description}

\item[\Cref{app:history}: Historical evolution, 2016--2026.] The trajectory from explicitly encoded inductive bias, through language-mediated representation and prompting, to controlled ablation, contamination auditing, and renewed specialized and jointly pretrained architectures. Contains \cref{fig:timeline}.
\item[\Cref{app:audit}: Search protocol, coverage audit, and risk of bias.] Supports \cref{sec:methodology}. The reference search and its measured sensitivity, the validation of the screening rule against an independent rater, and the SRB-6 risk-of-bias instrument with its per-study ratings. Contains \cref{fig:prisma,fig:rob} and \cref{tab:search-protocol,tab:risk-of-bias}.
\item[\Cref{app:surveys}: Positioning relative to existing surveys.] The full comparison against every survey located by a dedicated search, with the organizing axis of each and what it does and does not do relative to the substitution question. Contains \cref{tab:related-surveys}.
\item[\Cref{app:debate}: The scale-versus-structure debate.] The longer argument this review adjudicates empirically, from the bitter lesson through inductive-bias and form/meaning positions to the neurosymbolic programme, and the one physical-science precedent that already measures the gap as a function of scale.
\item[\Cref{app:disagreements}: What the field actually disagrees about.] The four kinds of disagreement recurring in this literature (empirical, mechanistic, definitional, and evaluative), separated by source, because each requires a different form of resolution, together with the strongest located counter-evidence to this review's own working hypothesis.
\item[\Cref{app:boundary}: Regime-boundary cases.] The taxonomy's harder classifications: protein structure as a composed Level-7-into-Level-2 case, code as an untested boundary, and the multi-level tagging rule for papers occupying two regimes at once.
\item[\Cref{app:evidence}: Benchmark and confound audits, and extended modality evidence.] The benchmark table, the confound-control audit of the twelve most load-bearing comparisons, and the extended graph, vision, and adjacent-modality evidence condensed in \cref{sec:graph-findings,sec:vision-findings,sec:adjacent-findings}. Contains \cref{tab:benchmarks,tab:confound-audit}.
\item[\Cref{app:theory}: Extended theoretical results.] The dispute over what mechanism in-context learning implements, the prompt-sensitivity literature, the pretraining-distribution results on task diversity and its trade-offs, and the scaling-law evidence whose exponents \cref{sec:scaling} declines to pool.
\item[\Cref{app:synthesis-tables}: Synthesis instruments.] The metric hierarchy, the mechanism-organized failure modes, the ten-claim map with its net synthesis judgment, and the open-problem restatement. Contains \cref{tab:metrics,tab:failure-modes,tab:claim-map,tab:open-problems} and \cref{fig:claimbalance}.
\item[\Cref{app:design}: Research questions and the minimal experimental design.] Each research question with its competing hypotheses, critical experiment, falsification condition, and confounds; four narrower and independently valuable directions; and the full specification of the proposed experiment.
\end{description}

\clearpage

\section{Historical Evolution, 2016--2026}
\label{app:history}

This appendix expands the trajectory summarized in \cref{sec:related-surveys}. It traces how a single question became progressively more precise: from whether structured data can be expressed through language, to whether language models perform competitively on such representations, to why they succeed or fail, and finally to what is transferred, preserved, computed, and learned through a language interface.

\paragraph{2016--2021: structure in the computation graph.} The dominant strategy was to encode known structure directly into the computation graph, because doing so remained one of the most reliable ways to generalize from limited domain-specific data. Group-equivariant convolutions incorporated symmetry into visual representations \citep{cohen2016group}; permutation-invariant architectures were developed for sets, from Deep Sets \citep{zaheer2017deepsets} to the attention-based Set Transformer \citep{lee2019set}; and message-passing networks for graphs \citep{kipf2017gcn,velickovic2018gat,xu2019gin} were unified under a framework of relational inductive biases \citep{battaglia2018relational}. In parallel, Perceiver IO \citep{jaegle2021perceiver} and Gato \citep{reed2022generalist} pursued generality through architecture rather than language, showing that the aspiration toward general-purpose models predates and does not require a language-mediated interface. What these approaches could not readily provide was transfer of a learned solution across modalities.

\paragraph{2022--2023: reduction to text.} From roughly 2022, researchers investigated whether specialization could be reduced by representing structured objects as text and prompting or lightly fine-tuning pretrained language models: TabLLM \citep{hegselmann2023tabllm}, LLMTime \citep{gruver2023large}, and early graph-verbalization approaches \citep{wang2023nlgraph,fatemi2024talk}. The optimism of this period is reflected in position work arguing that graph foundation models were already emerging as a paradigm \citep{mao2024position}. What many early results did not establish was whether the language model itself was responsible for the observed gains.

\paragraph{2024--2025: the evidentiary standard changes.} Controlled ablations increasingly tested whether the language-model component was load-bearing: removing or reinitializing it in three widely cited time-series forecasters did not reduce performance \citep{tan2024are}, and contamination audits examined whether apparent few-shot advantages reflected pretraining exposure \citep{golchin2024time,bordt2024elephants}. Theory supplied a complementary account, with \citet{kozachinskiy2025strassen} establishing a precision-independent ceiling on the compositional operations achievable by a single attention layer of sub-polynomial size. Together these changed what counts as evidence: high benchmark accuracy ceased to be sufficient grounds for concluding that a language model had acquired the structure a task requires.

\paragraph{2025--2026: specialization scales, and joint pretraining appears.} Specialized non-linguistic architectures continued to scale independently \citep{grinsztajn2025tabpfn25,qu2026tabiclv2,becherler2026billion,nxai2025tirex}, and a different strategy emerged: rather than adapting a pretrained language model after the fact, Chronicle jointly pretrains language and time-series representations through shared attention blocks \citep{quinlan2026chronicle}. Meanwhile the cases in which language-mediated systems provide clear practical value increasingly involve orchestration rather than replacement (\cref{sec:autoML}).

\section{Search Protocol, Coverage Audit, and Risk-of-Bias Assessment}
\label{app:audit}
\label{sec:coverage-audit}

The search described in \cref{sec:methodology} was structured but not preregistered, and was carried out by a single author. Three instruments are reported here so that a reader can judge what that cost: a reproducible reference search with a measured sensitivity (\cref{app:refsearch}), a validation of the screening rule against an independent reference rater (\cref{app:screening}), and a risk-of-bias assessment of the primary studies this review's conclusions rest on (\cref{app:rob}). None of the three converts this review into a systematic review, and it is not presented as one. The protocol, including its registration status and its limitations, is released with the manuscript.

\subsection{Reference search}
\label{app:refsearch}

\begin{figure}[t!]
\centering
\includegraphics[width=0.92\textwidth]{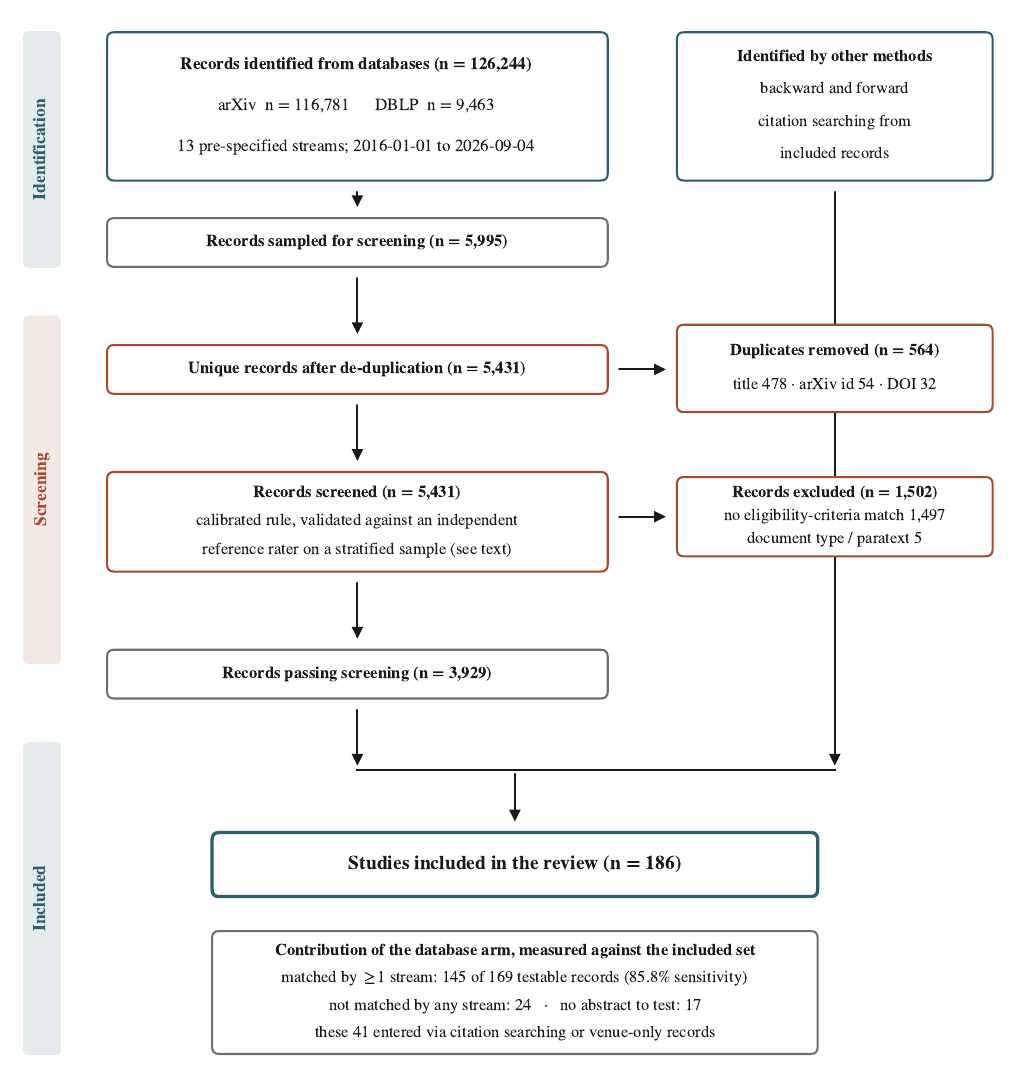}
\caption{PRISMA-style flow for the reference search described in \cref{app:refsearch}. The identified set is too large to retrieve in full, so each stream was sampled by drawing random page offsets across its full result range; screening counts are therefore sample counts, and the rates they imply generalize to the identified set. The bottom panel reports what the database arm contributed to the included set.}
\label{fig:prisma}
\end{figure}

A reference search was specified in code, issued to arXiv and DBLP, and is reported in PRISMA-style flow in \cref{fig:prisma}; the thirteen stream expressions and their per-stream yields appear in \cref{tab:search-protocol}. Two databases were used because neither suffices alone: arXiv, because 40\% of this review's corpus is arXiv-only preprint literature, and DBLP for peer-reviewed computer-science venues. OpenAlex was attempted first and abandoned when its rate limiter blocked bulk retrieval partway through; this is recorded because it bounds the identification counts.

The search identified 126,244 records (116,781 from arXiv, 9,463 from DBLP). Retrieving all of them was not feasible, so each stream was sampled: random page offsets were drawn uniformly across the stream's full result range and every record in the drawn page was taken, giving an equal-probability cluster sample of 5,995 records. arXiv's paging limit exceeds every stream's count, so no stream is truncated, and the offsets are drawn from a seeded generator so the sample is reproducible from the protocol. Screening counts are therefore sample counts, but the rates they imply generalize to the identified set, which a relevance-ordered head would not. After removing 564 duplicates, 5,431 unique records remained, of which 3,929 passed screening.

The strategy was developed against a seed set rather than fixed in advance. Successive versions were tested offline against the title and abstract text of the corpus's own records and revised wherever a version failed to retrieve records already known to be relevant. Two versions were discarded, and the protocol reports both: one too specific to recover the foundational architecture literature at all, and one sensitive enough but identifying roughly 865{,}000 records, which is not screenable by any procedure available here. The adopted strategy recovers 145 of the 169 corpus records that have a retrievable abstract, a sensitivity of 85.8\%.

That figure is the substantive result of the audit, and it does not flatter the method. A developed boolean strategy over two databases still misses roughly one known-relevant record in seven, and the misses are systematic: they concentrate in foundational architecture papers whose abstracts never use language-model vocabulary, and in papers that avoid the field's standard terms altogether. Both groups entered this corpus through backward and forward citation tracing, which PRISMA~2020 treats as a separate identification arm precisely because it cannot be reproduced from a query string. A further 17 records have no arXiv abstract against which the strategy could be tested. Together these 41 records measure what a database-only search would have missed, and they are the reason this review is reported as a narrative synthesis.

\begin{table}[t!]
\centering

\caption{The thirteen pre-specified search streams, with the Boolean expression issued to each database and the number of records each returned (\cref{sec:methodology}). Each stream is a conjunction of term groups: a learning-system group, a modality or problem group, and, for most streams, a structural-question group; the same expression was compiled into the native query syntax of both databases so that the two searches are identical in content.}

\label{tab:search-protocol}

\adjustbox{max width=\textwidth,max totalheight=0.80\textheight,center}{%
\begin{tabular}{@{}
    >{\centering\arraybackslash}p{0.045\textwidth}
    >{\raggedright\arraybackslash}p{0.10\textwidth}
    >{\raggedright\arraybackslash\arraybackslash}p{0.63\textwidth}
    >{\centering\arraybackslash}p{0.075\textwidth}
    >{\centering\arraybackslash}p{0.075\textwidth}
    >{\centering\arraybackslash}p{0.075\textwidth}
@{}}

\toprule
\textbf{ID} &
\textbf{Stream} &
\textbf{Boolean expression} &
\textbf{arXiv} &
\textbf{DBLP} &
\textbf{Total} \\
\midrule

S1 &
tabular &
\texttt{(large language model OR language model OR foundation model OR in-context learning OR llm) AND (tabular OR gradient-boosted OR feature engineering OR xgboost)} &
1,564 & 463 & 2,027 \\

S2 &
graph &
\texttt{(large language model OR language model OR foundation model OR in-context learning OR llm) AND (graph neural network OR graph reasoning OR graph learning OR gnn OR knowledge graph OR node classification OR graph structure)} &
3,940 & 944 & 4,884 \\

S3 &
time series &
\texttt{(large language model OR language model OR foundation model OR in-context learning OR llm) AND (time series OR forecasting)} &
2,397 & 1,313 & 3,710 \\

S4 &
vision &
\texttt{(vision-language model OR multimodal large language model OR vision language model OR visual language model OR multimodal llm OR large multimodal model) AND (spatial OR visual reasoning OR compositional OR visio-linguistic OR shortcoming OR perceive OR benchmark OR geometry)} &
10,365 & 723 & 11,088 \\

S5 &
icl-theory &
\texttt{(in-context learning OR in-context) AND (inductive bias OR sample complexity OR task diversity OR gradient descent OR bayesian OR function class OR phase transition OR pretraining distribution OR linear regression OR induction head OR algorithm)} &
21,779 & 23 & 21,802 \\

S6 &
expressivity &
\texttt{(transformer OR attention mechanism OR graph neural network OR neural network) AND (expressive power OR vc dimension OR universal approximation OR expressivity OR impossibility OR rademacher OR sample complexity OR lower bound OR compositionality OR scaling law OR positional encoding)} &
21,447 & 80 & 21,527 \\

S7 &
chemistry &
\texttt{(large language model OR language model OR foundation model OR in-context learning OR llm OR smiles OR molecular representation) AND (molecular property OR chemistry OR drug discovery OR molecular graph OR molecule)} &
1,567 & 203 & 1,770 \\

S8 &
code &
\texttt{(large language model OR language model OR foundation model OR in-context learning OR llm) AND (code execution OR program semantics OR code reasoning OR code generation OR execution OR python function)} &
12,607 & 170 & 12,777 \\

S9 &
point cloud &
\texttt{(point cloud OR 3d object OR 3d shape OR 3d understanding) AND (language model OR multimodal OR captioning OR spatial OR encoder OR llm)} &
4,286 & 374 & 4,660 \\

S10 &
protein &
\texttt{(protein language model OR protein structure OR amino acid) AND (structure-aware OR structural OR sequence OR fitness OR contrastive)} &
2,445 & 989 & 3,434 \\

S11 &
evaluation &
\texttt{(data contamination OR benchmark contamination OR train-test leakage OR memorization OR data leakage OR output format OR prompt order OR long context OR prompt sensitivity) AND (language model OR pretraining data OR benchmark OR generalization OR llm)} &
6,349 & 130 & 6,479 \\

S12 &
architecture &
\texttt{(equivariant OR equivariance OR permutation invariant OR permutation-invariant OR relational inductive bias OR geometric deep learning OR message passing OR inductive bias OR convolutional network OR set transformer OR deep sets) AND (neural network OR architecture OR graph OR set OR point cloud OR image OR time series OR generalization OR symmetry OR structure)} &
20,913 & 781 & 21,694 \\

S13 &
surveys &
\texttt{(survey OR systematic review OR position paper OR opportunities and risks OR foundations and trends) AND (large language model OR foundation model OR graph OR tabular OR time series OR in-context learning OR multimodal OR protein OR point cloud OR augmented)} &
7,122 & 3,270 & 10,392 \\

\midrule

\multicolumn{3}{@{}l}{\textbf{Total records identified}} &
\textbf{116,781} &
\textbf{9,463} &
\textbf{126,244} \\

\bottomrule

\end{tabular}%
}

\end{table}

\subsection{Screening validation}
\label{app:screening}

With a single author there is no second human screener. The substitute adopted here is to treat the screening rule as an instrument and validate it against an independent reference rater, which is the form of dual screening that can actually be checked in this situation. The rule was applied to the 2,710 pooled records carrying an abstract, and a validation sample was drawn \emph{stratified by the rule's own output}: 120 records from the 1,870 it included and 120 from the 840 it excluded, presented in random order. A reference rater then applied the written eligibility criteria to title and abstract, blind to the rule's labels and to its definitions.

Stratification is what makes the resulting statistics interpretable. Screening a pool drawn from the head of a relevance ranking yields degenerate marginals: both raters include nearly everything, chance agreement approaches observed agreement, and $\kappa$ collapses for reasons that concern the sampling frame, not the raters. Drawing the pool as a probability sample and then balancing the validation strata holds expected agreement at 50.0\% by design. Observed agreement is 76.7\%, giving $\kappa = 0.53$ (95\% CI 0.43--0.64), moderate on the Landis--Koch scale.

For a screening instrument, however, agreement is not the quantity that matters most; the asymmetry of the errors is. Passing an ineligible record forward costs one full-text assessment, whereas discarding an eligible one removes it from the review permanently. The rule's sensitivity against the reference rater is 87.2\% (95\% CI 78.5--92.7) within the sample and its specificity 70.8\% (95\% CI 63.2--77.4); reweighting the strata back to their true sizes gives a population sensitivity of 93.8\% against a specificity of 52.1\%. The rule is therefore deliberately over-inclusive, as a screen should be: it retains about nine in ten eligible records and passes a substantial number of ineligible ones to the next stage, where full-text assessment removes them. The honest cost is the remainder: an estimated 77 eligible records, 6.2\% of those in the pool, are discarded at screening and would not be recovered without the citation-tracing arm.

\subsection{Risk of bias}
\label{app:rob}

\begin{figure}[t!]
\centering
\includegraphics[width=\textwidth]{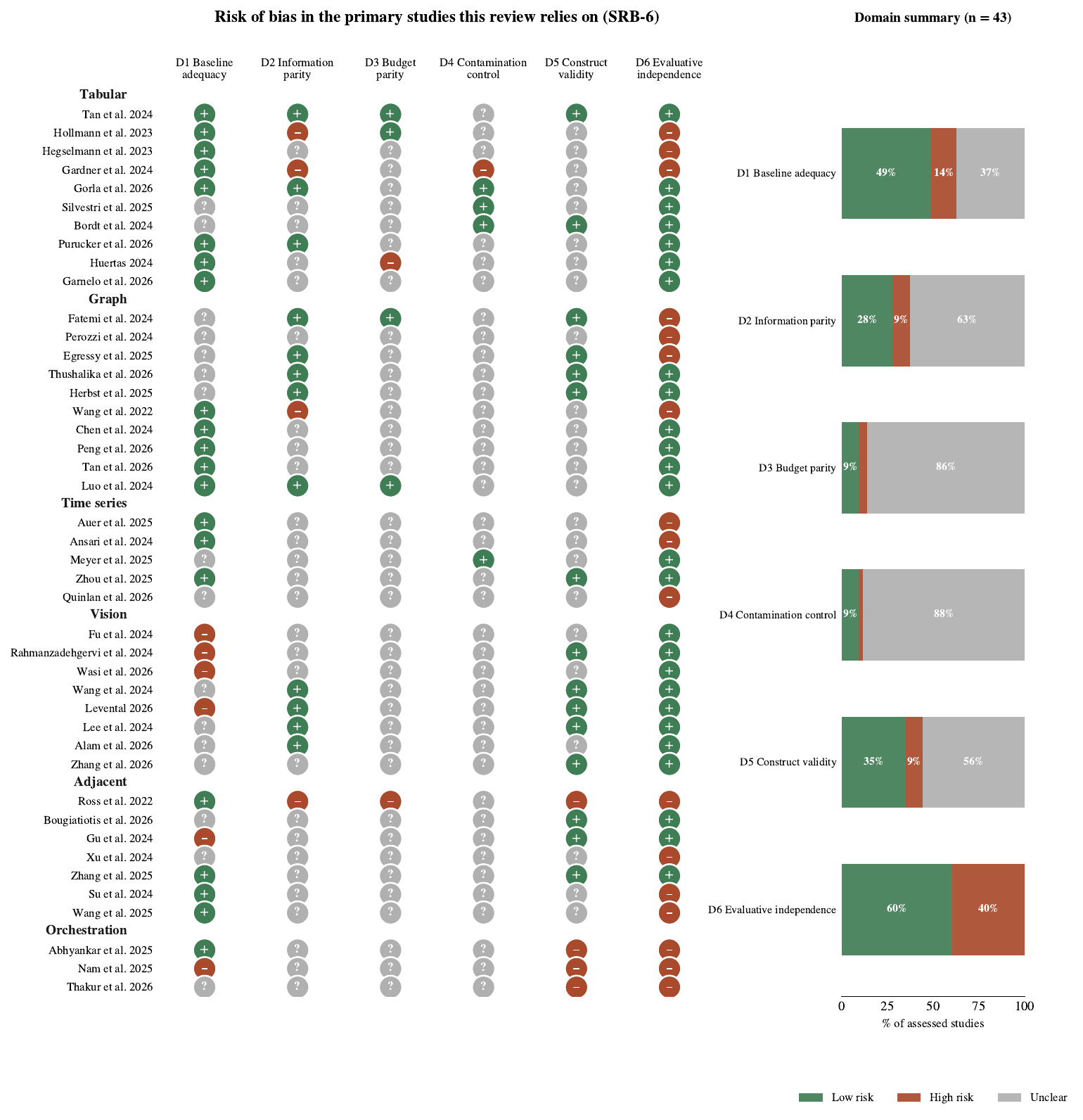}
\caption{Risk of bias across the primary studies whose empirical results this review relies on, assessed with the SRB-6 instrument defined in \cref{tab:risk-of-bias}. Left: per-study ratings, grouped by modality. Right: the distribution within each domain. \emph{Unclear} records that an item could not be established from the sources read for this review, and is a statement about reporting practice rather than about study quality; it is the modal rating in four of the six domains.}
\label{fig:rob}
\end{figure}

Clinical instruments such as ROBIS and Cochrane RoB presuppose a treatment-effect design and do not transfer to comparisons between learning systems. \Cref{tab:risk-of-bias} therefore defines a domain-specific instrument, SRB-6, derived from the three confounds already coded in \cref{tab:confound-audit} and extended with three domains the evidence in \cref{sec:methods} showed to be load-bearing: baseline adequacy, information parity, budget parity, contamination control, construct validity, and evaluative independence. Each of the 43 primary studies whose empirical results this review relies on was rated \emph{low}, \emph{high}, or \emph{unclear} on every domain (\cref{fig:rob}). Formal results are excluded by design, since every domain presupposes an empirical comparison.

The distribution is itself a finding, and it is the quantitative form of the argument \cref{sec:comparison-scientific} makes case by case. Of the 258 ratings, 82 (32\%) are low risk, 34 (13\%) are high risk, and 142 (55\%) are \emph{unclear}, the item could not be established from the source at all. The two domains that matter most for a replacement claim are the worst served: 86\% of studies do not report enough to judge budget parity, and 88\% do not address contamination. 40\% are evaluated only by the authors of the system being evaluated. Exactly three studies are low risk on four or more domains with no high-risk rating: \citet{tan2024are}, \citet{gorla2026illusion}, and \citet{luo2024classic}. That all three are ablations or independent re-analyses, and none is a system paper, is consistent with this review's central methodological claim, the evidence that survives scrutiny is the evidence that was designed to be scrutinised.

Two limitations bound the instrument. Ratings were made by a single assessor, so no agreement statistic is available for it; and \emph{unclear} throughout is a statement about reporting practice, not about study quality. A study rated unclear on contamination has not been shown to be contaminated, only to be silent on the question.

\begin{table}[t]
\centering
\small

\caption{The SRB-6 risk-of-bias instrument and its domain-level results (\cref{sec:coverage-audit}, \cref{fig:rob}). Clinical instruments such as ROBIS and Cochrane RoB assume a treatment-effect design and do not transfer to comparisons of learning systems, so the six domains below were derived from the three confounds already coded in \cref{tab:confound-audit} and extended with three further domains that the evidence in \cref{sec:methods} showed to be load-bearing. Each of the 43 primary studies whose empirical results this review relies on was rated \emph{low}, \emph{high}, or \emph{unclear} on every domain. \emph{Unclear} records that the item could not be established from the sources read for this review; it carries no judgement about the study. Formal results are excluded by design, since every domain presupposes an empirical comparison.}

\label{tab:risk-of-bias}

\begin{tabular}{@{}l l p{0.34\textwidth} p{0.40\textwidth}@{}}
\toprule
\textbf{ID} &
\textbf{Domain} &
\textbf{Question} &
\textbf{Criterion for \emph{high} risk} \\
\midrule

D1 &
Baseline adequacy &
Is a matched, tuned, non-linguistic specialized baseline reported for the claim being made? &
\emph{High} if the only comparators are human performance, other language-mediated systems, or none. \\[0.35em]

D2 &
Information parity &
Do the compared arms have equal access to task-relevant information? &
\emph{High} where one arm receives textual side-information or domain pretraining the other cannot use. \\[0.35em]

D3 &
Budget parity &
Are compute, parameter, or data budgets matched, or at minimum reported? &
\emph{High} where a reported accuracy parity conceals an order-of-magnitude cost difference. \\[0.35em]

D4 &
Contamination control &
Is overlap between the evaluation set and pretraining data addressed? &
\emph{High} where a subsequent audit established contamination the original study did not detect. \\[0.35em]

D5 &
Construct validity &
Does the reported metric test the structural property claimed, rather than an accuracy proxy? &
\emph{High} where a structural or mechanistic claim rests solely on end-task accuracy. \\[0.35em]

D6 &
Evaluative independence &
Is the evidence produced by parties other than those proposing the method? &
\emph{High} where the only evaluation of a system is by its own authors. \\

\bottomrule
\end{tabular}

\vspace{0.9em}

\begin{tabular}{@{}p{0.55\textwidth} r r r@{}}
\toprule
\textbf{Domain} &
\textbf{Low risk} &
\textbf{High risk} &
\textbf{Unclear} \\
\midrule

D1 Baseline adequacy &
21 (49\%) &
6 (14\%) &
16 (37\%) \\

D2 Information parity &
12 (28\%) &
4 (9\%) &
27 (63\%) \\

D3 Budget parity &
4 (9\%) &
2 (5\%) &
37 (86\%) \\

D4 Contamination control &
4 (9\%) &
1 (2\%) &
38 (88\%) \\

D5 Construct validity &
15 (35\%) &
4 (9\%) &
24 (56\%) \\

D6 Evaluative independence &
26 (60\%) &
17 (40\%) &
0 (0\%) \\

\midrule

\textbf{All domains pooled} (258 ratings) &
82 (32\%) &
34 (13\%) &
142 (55\%) \\

\bottomrule
\end{tabular}

\end{table}

\section{Positioning Relative to Existing Surveys}
\label{app:surveys}

This appendix expands \cref{sec:related-surveys}. \Cref{tab:related-surveys} sets out every survey located by a dedicated search, the axis on which each is organized, whether it addresses the substitution question, and the depth to which each was read for this review. The discussion below treats the entries that require differentiation rather than mere listing.

\begin{table}[t!]
\centering
\setlength{\tabcolsep}{4pt}
\renewcommand{\arraystretch}{0.92}

\caption{Key existing surveys and reviews most closely related to this review,
showing their scope and the principal gap addressed here.}
\label{tab:related-surveys}

\begin{tabularx}{\textwidth}{@{}
    >{\raggedright\arraybackslash}p{0.19\textwidth}
    >{\raggedright\arraybackslash}p{0.23\textwidth}
    >{\raggedright\arraybackslash}X
@{}}
\toprule
\textbf{Survey} &
\textbf{Scope} &
\textbf{Gap relative to this review} \\
\midrule

\citet{liang2024foundations}
(peer-reviewed, ACM \emph{Computing Surveys})
&
Multimodal ML broadly
&
Strongest structural precedent: separates \emph{representation} from
\emph{reasoning} across modalities. This review additionally distinguishes
compute from learn and develops an eight-sense replacement framework
(\cref{tab:replacement}). \\

\citet{sun2026survey}
(full text read)
&
Tabular, time series, graph; 150+ methods
&
Closest overall competitor. Shares much of the primary literature and
identifies permutation invariance as a core bias, but does not test whether
LLM-based methods satisfy it or use controlled ablations and contamination
re-analysis to evaluate structural substitution. \\

\citet{zhang2026suitable}
&
Graph computation only
&
Examines LLM-as-executor versus LLM-as-planner, overlapping with this
review's reasoning--orchestration boundary (\cref{sec:regime-ladder}).
Does not address GNN replacement, permutation invariance, or
cross-modality synthesis. \\

\citet{jin2026large}
(peer-reviewed, ACM \emph{Computing Surveys})
&
Time series and spatio-temporal data
&
The closest modality-level competitor located. Provides a method
taxonomy but does not frame temporal structure as an architectural bias that
may or may not be substituted by language-mediated computation. \\

\citet{liu2025graph}
&
Graph foundation models
&
Peer-reviewed TPAMI survey covering GFMs on their own architectural terms,
with no comparison of language-mediated approaches against specialized
structural inductive biases. \\

\citet{liu2025multimodal}
&
Time series with other modalities
&
Closest existing analogue to the review's central axis: \emph{TimeAsX}
treats converting structured signals into another representation as a
first-class category. It remains confined to time series and does not
distinguish what systems describe, preserve, compute, or learn. \\

\bottomrule
\end{tabularx}

\end{table}

\subsection{Modality-specific surveys}

The literature contains substantial surveys of LLMs for individual modalities: tabular data \citep{fang2024large,wu2025tabular}, graphs \citep{jin2023large,ren2024survey}, time series \citep{zhang2024large,jin2026large}, chemistry \citep{han2025from}, protein sequences \citep{xiao2025protein}, point clouds \citep{thengane2025foundational}, tabular data via graph-structured representations \citep{li2025graph}, and multimodal LLMs for scientific discovery more broadly \citep{s3bench2025comprehensive}. These organize the literature by modality-specific architectures, representations, adaptation strategies, or downstream tasks; none organizes it by what a system can be shown to compute.

\citet{jin2026large} is the closest modality-level competitor located, published roughly one month before this manuscript's arXiv posting. It organizes time-series and spatio-temporal large models along data type, model category, model scope, and application area, but does not frame temporal locality or shift-invariance as a substitutable inductive bias in the sense of \cref{sec:distinction}. \citet{li2025graph} surveys the reverse direction of this review's tabular/graph relationship: relocating graph-structural bias \emph{into} tabular learning through GNNs rather than through language. That is a structural analogue of this review's relocation thesis, argued from the specialized-architecture side.

The graph-specific surveys \citep{liu2025graph,wang2025graphfoundation} treat graph foundation models on their own architectural terms without a language-substitution comparison. \citet{yu2025graph2text} organizes the graph--LLM literature specifically along a graph-to-text-versus-graph-to-token axis that is, within graphs alone, the same Level-2/Level-7 boundary this review treats as load-bearing across nine modalities (\cref{sec:regime-ladder}); it does not connect that boundary to theory, to a replacement taxonomy, or to any modality beyond graphs. \citet{zhang2026suitable} independently reaches this review's reasoning-versus-orchestration boundary through an LLM-as-executor-versus-LLM-as-planner taxonomy for graph computation, reporting an overlapping conclusion (reliable for small-scale tasks, unreliable at scale) without addressing GNN replacement, permutation invariance, or synthesis across modalities. Its independent convergence on the same boundary is treated here as corroboration, not as prior claim to it.

\subsection{Cross-modal and structural precedents}

\citet{sun2026survey} reviews LLM-based methods for tabular data, time series, and graphs, providing a broader cross-modal engineering taxonomy based on data representation, model architecture, pretraining, and adaptation. \citeauthor{sun2026survey} observe that reducing a task to a language-mediated representation often loses explicit structural or relational inductive bias, but do not test that claim directly.

The closest existing analogue to this review's central axis is \citet{liu2025multimodal}, whose \emph{TimeAsX} category names precisely the move this review interrogates -- reusing a foundation model pretrained on another modality to analyze structured data -- alongside \emph{Time+X} for multimodal input and \emph{Time2X/X2Time} for cross-modal translation. That taxonomy classifies what a system is built to do, and it is confined to time series; it does not ask whether the borrowed model preserves or computes the structure the task requires.

\citet{liang2024foundations} is the closest structural precedent located overall. It organizes a fragmented multimodal-ML literature around six technical challenges -- \emph{representation} and \emph{reasoning} among them, as explicitly separate challenges -- and three cross-cutting principles, in place of modality or construction pattern. This review's increment relative to Liang et al.\ is the further compute/learn split within what they treat as a single reasoning challenge, and the eight-sense decomposition of ``replacement'' (\cref{tab:replacement}) that neither Liang et al.\ nor any other survey located here proposes.

\citet{mahowald2024dissociating}'s formal-versus-functional-competence distinction is the closest interdisciplinary analogue to this review's central framework: a comparable ``knowing versus doing'' structure, applied to linguistic competence in place of structured non-linguistic data. Related surveys of in-context learning \citep{dong2024survey,zhou2024mystery,mao2025survey} and geometric deep learning \citep{bronstein2021geometric} address complementary aspects of the problem but do not connect architectural inductive bias, representation, computation, and learning within a common framework. \citet{kostikova2026lllms} maps the literature on LLM limitations data-driven\-ly and at corpus scale; it establishes \emph{that} limitations are widely reported without asking, for any structured modality, whether a reported limitation reflects representation, computation, or evaluation, which is the separation this review supplies. Two broader position pieces set the context within which all of these sit: \citet{bommasani2022opportunities} names the foundation-model paradigm and catalogues its capabilities and risks, and \citet{ye2025llms4all} surveys LLM applications organized by academic discipline. Neither proposes a substitution-specific analytical framework, and neither is structured around what a system can be shown to compute.

\subsection{Summary of the gap}

No survey located in this search spans more than two or three of this review's nine structured-data modalities under one consistent framework, and none proposes an analogue of the four-way describe/preserve/compute/learn distinction or the eight-sense replacement decomposition. The contribution of this review is therefore the use of that common framework to synthesize evidence across modalities, connect empirical findings with learning theory, and identify the controlled experimental gap that remains unresolved (\cref{sec:distinction,sec:evaluation,sec:open-problems}).

\section{The Scale-versus-Structure Debate This Review Adjudicates}
\label{app:debate}
\label{sec:scale-vs-structure}

The question asked here is not new. \citet{sutton2019bitter} argues from seventy years of AI research that general methods scaling with computation have consistently outperformed hand-encoded domain structure. \citet{goyalbengio2022inductive} argue the converse: reliable out-of-distribution and systematic generalization requires inductive biases that generic scale does not supply. \citet{bender2020climbing} make a related but distinct claim about language specifically: a system trained only on form has no way in principle to acquire meaning, so fluency is not evidence of the understanding it appears to display. \citet{mahowald2024dissociating} operationalize a version of this for linguistic competence, separating formal facility with language from functional competence at the reasoning tasks language describes. And the neurosymbolic programme has argued for hybrid architectures at least since \citet{marcus2020next}: when a purely neural approach fails, an explicit structural component is reattached.

The same argument is furthest along empirically in the physical sciences. \citet{ngo2025scaling} fit scaling laws for equivariant and non-equivariant neural force fields and report that the scaling exponent is architecture-dependent, with symmetry-respecting architectures retaining their advantage as scale grows -- a $\Delta(N)$-style measurement of exactly the kind \cref{sec:scaling} finds missing for language-mediated models, obtained where the relevant symmetry group is unambiguous.

This review is best understood as a systematic empirical adjudication of that debate, restricted to language-mediated systems and structured non-linguistic data. Most of the debate has been conducted by assertion and isolated demonstration: Sutton's argument is historical; Goyal and Bengio's is a list of candidate biases with no corpus-level test of whether current systems possess them; Bender and Koller's is conceptual rather than evidentiary; and the neurosymbolic literature's instances are typically demonstrations offered by researchers already committed to the reattachment thesis. What distinguishes the three convergent cases in \cref{sec:introduction} is that none of the communities involved set out to argue for symbolic reattachment, cite the neurosymbolic literature, or share authorship. The four-way distinction of \cref{sec:distinction} should accordingly be read as an operationalization of the Bender-and-Koller and Mahowald-et-al.\ form/competence arguments, turned into a per-modality, per-claim evidentiary test and applied to structured data in place of language itself.

\section{What the Field Actually Disagrees About}
\label{app:disagreements}
\label{sec:problem-formulation}

Some disagreements in this literature are genuinely empirical; others arise because different studies answer different questions in similar language. Underlying most of them is a single habit: success at one stage of the learning process is routinely read as evidence for success at another. \citet{vafa2025what}'s inductive-bias probe illustrates why that inference fails, a transformer achieves near-perfect accuracy on orbital-trajectory prediction while failing to internalize the underlying physical structure, breaking when evaluated on computationally different but physically equivalent targets. High task accuracy can therefore coexist with failure to recover the structure that would make that accuracy robust, which is the observation \cref{sec:distinction} formalizes. Four disagreements recur, and they are distinguished here by their source, because each requires a different form of resolution. A disagreement is \emph{empirical} when competing claims rest on measured results not reconciled under shared conditions; \emph{mechanistic} when researchers agree on what a system achieves but disagree about why; \emph{definitional} when one term carries two substantive claims; and \emph{evaluative} when the dispute is whether a reported result would survive scrutiny not yet applied.

\emph{The empirical disagreement} concerns the scope of the substitution claim. One line advances it strongly: \citet{song2024omnipred} present language models as universal regressors, \citet{wang2023unipredict} as universal tabular classifiers. A second line tests it under controlled conditions and finds important qualifications. Removing the language-model component from three widely cited time-series forecasters does not reduce performance \citep{tan2024are}; changing only the textual encoding of a fixed graph task produces accuracy swings of 4.8--61.8 percentage points \citep{fatemi2024talk}; and the causal attention mask alone lets a decoder-only transformer recover absolute token position even with no positional encoding \citep{kazemnejad2023impact}, so such models are not permutation-invariant whatever they can describe \citep{egressy2025setllm}. Both sides can point to measured results, established under different experimental conditions.

\emph{The mechanistic disagreement} concerns whether structural constraints must be embedded in the architecture itself. The working hypothesis adopted throughout this review is that structural assumptions generally need to be reflected in the computation rather than merely supplied at inference time, and that position is not uncontested. \citet{mittal2025architectural} show that a standard causally masked transformer, given an appropriate \emph{inferential} bias (multi-step autoregressive generation rather than single-step prediction), can match and sometimes outperform custom permutation-invariant architectures on exchangeable sequence modeling, and report that those custom architectures do not themselves guarantee exchangeability. This is the most direct counter-evidence located to the stronger form of that hypothesis: for at least some structural properties, the architectural-versus-inferential distinction may matter more than the architectural-versus-language-mediated one.

\emph{The definitional disagreement} concerns the word ``replace.'' Work on LLM-driven feature engineering, model search, and pipeline construction \citep{abhyankar2025llmfe,nam2025mlestar}, and position papers arguing that graph foundation models are already here \citep{mao2024position}, use it to mean that a language model now performs a role previously carried out by a human or a specialized system. This review uses it to mean that the language model's own computation implements the structural function rather than delegating it. Neither usage is privileged here; \cref{tab:replacement} separates eight senses precisely because many apparent disagreements turn out to be disagreements about which claim is being made.

\emph{The evaluative disagreement} has, in one prominent case, already been settled by direct re-examination. Tabula-8B was originally reported to outperform XGBoost and TabPFN by 5--15 percentage points in few-shot tabular prediction \citep{gardner2024tabula8b}. A re-analysis found that instruction-tuning the same base model with no tabular exposure reaches 92.2\% of Tabula-8B's own mean classification accuracy, and identified train--test contamination that standard deduplication had failed to detect \citep{gorla2026illusion,shi2024detecting}. That 92.2\% is an absolute-accuracy comparison, not a recovery of the specific margin originally reported. Similar concerns arise elsewhere \citep{silvestri2026when,bordt2024elephants}, but comparable audits have not been performed across most benchmarks in \cref{tab:benchmarks}. How much positive evidence would survive equivalent scrutiny remains open.

\section{Regime-Boundary Cases}
\label{app:boundary}

This appendix expands \cref{sec:regime-ladder}. The eight regimes are stated there as a clean partition; in practice several reviewed systems sit on a boundary, and how those cases are read affects what the taxonomy is claimed to do. Three classes of boundary case are treated here: papers occupying two regimes at once, compositional cases in which one regime's output becomes another's input, and one modality for which the boundary is untested rather than confirmed.

\subsection{Papers occupying two regimes}

The regimes are not mutually exclusive at the level of a paper, because the unit classified throughout this review is the reported result rather than the publication (\cref{sec:methodology}). \Cref{tab:method-comparison} therefore tags several methods at two levels simultaneously, and does so because the underlying results genuinely occupy two regimes. TabLLM \citep{hegselmann2023tabllm} reports both a plain-serialization baseline and a few-shot demonstration condition, placing it at Levels 2 and 4. SimKGC's textual scoring model sits at the same boundary \citep{wang2022simkgc}. LLM-FE \citep{abhyankar2025llmfe} both generates code and searches with it, placing it at Levels 5 and 6. A single method can legitimately occupy different cells for different components or conditions, and forcing a single label would discard the distinction the taxonomy exists to draw.

\subsection{Time-LLM: a name that does not match a mechanism}

The most informative boundary case is Time-LLM \citep{jin2024timellm}, marketed and named around ``reprogramming'' time series through language. \Cref{tab:method-comparison} classifies its actual mechanism at Level 7, not Levels 1--4, because the structural computation it performs on the input sequence never runs through interpretable language tokens. The ablation evidence reviewed in \cref{sec:timeseries-findings} matches what that classification predicts: removing the language-model component does not hurt performance \citep{tan2024are}. Were the mechanism genuinely Level 1--4, removing the language component should remove the computation itself and leave nothing intact. This is the sense in which the taxonomy is explanatory rather than merely descriptive: it generates a testable expectation before the ablation is run.

\subsection{Protein structure: a Level 7 computation feeding a Level 2 representation}

Protein structure produces a genuine compositional case that the eight discrete levels do not separate cleanly, and should not be forced to. \citet{su2024saprot} injects Foldseek-derived structure tokens -- the output of a separate, non-linguistic structural-alignment tool -- directly into a sequence model's vocabulary, so the structural computation originates at Level 7 but is consumed as ordinary sequence tokens at Level 2. \citet{wang2025splm} reaches the same destination by a different route, aligning a sequence encoder with a separate structure encoder through contrastive learning so that structural information is distilled into a sequence-only representation available at inference. The structural computation again happens outside the sequence model, but here it is absorbed into the encoder's weights rather than its vocabulary.

Read compositionally -- a Level 7 computation feeding a Level 2 representation -- both cases are handled by the same rule already applied to TabLLM and SimKGC above. What makes them worth stating separately is that the composition happens inside the vocabulary or the weights rather than across visibly separate components, which is exactly the configuration in which a system's name stops advertising where its structural computation lives. This is the taxonomic form of the relocation argument in \cref{sec:replacement-definitions}.

\subsection{Point clouds and knowledge graphs: the boundary reproduced directly}

Two adjacent modalities reproduce the boundary without complication. PointLLM pairs a specialized point-cloud encoder with a language interface, placing the structural representation at Level 7 while language handles only captioning and interaction \citep{xu2024pointllm}. Knowledge-graph agents with extensive tool access remain at Level 6 regardless of that access, and continue to underperform specialized graph models at Level 8 \citep{tan2026gabench}. Neither case requires a compositional reading.

\subsection{Code: an untested boundary}

Code is the weakest fit of the nine modalities. The reviewed evidence \citep{gu2024cruxeval} concerns whether models correctly simulate program execution, which is a describe-versus-compute question in the sense of \cref{sec:distinction} rather than a regime-boundary one, and no non-linguistic Level 8 baseline is reported for comparison. For code, the regime boundary is therefore \emph{untested} rather than confirmed, and it is reported as such throughout: \cref{sec:benchmarks} records the corresponding absence of a specialized baseline in CRUXEval.

\section{Benchmark and Confound Audits, and Extended Modality Evidence}
\label{app:evidence}
\label{app:modality-detail}

This appendix holds the three evidence tables the main text refers to but does not display, together with the modality-level detail condensed in \cref{sec:graph-findings,sec:vision-findings,sec:adjacent-findings}.

\subsection{Method classification}
\label{app:method-table}

\Cref{tab:method-comparison} records, for twenty representative methods across the four core modalities, the representational regime each occupies under \cref{fig:regimeladder} and its principal limitation. The classification is the per-system form of the argument \cref{sec:methods} makes modality by modality: the strongest systems in every modality sit at Levels 7--8, and every system whose headline framing suggests language-mediated computation but whose mechanism does not, Time-LLM most clearly, is classified by mechanism rather than by name (\cref{app:boundary}).

\begin{table}[t!]
\centering
\footnotesize
\setlength{\tabcolsep}{3.5pt}
\renewcommand{\arraystretch}{0.95}

\caption{Representative methods across the four core modalities, focusing on their representational regime and principal limitation. Representation levels follow the taxonomy in \cref{fig:regimeladder}.}

\label{tab:method-comparison}

\begin{lrbox}{\wtabbox}
\begin{tabular}{@{}
    >{\raggedright\arraybackslash}p{0.12\textwidth}
    >{\raggedright\arraybackslash}p{0.20\textwidth}
    >{\raggedright\arraybackslash}p{0.22\textwidth}
    >{\raggedright\arraybackslash}p{\dimexpr 0.46\textwidth-21pt\relax}
@{}}
\toprule
\textbf{Modality} &
\textbf{Method} &
\textbf{Representation} &
\textbf{Main limitation} \\
\midrule

\multirow{6}{*}{Tabular}
& TabPFN-2.5 \citep{grinsztajn2025tabpfn25}
& Numeric ICL (L8)
& Not language; limited to small--medium tables \\

& TabICLv2 \citep{qu2026tabiclv2}
& Numeric ICL (L8)
& Not language; primarily classification-focused \\

& TabLLM \citep{hegselmann2023tabllm}
& Text serialization (L2/4)
& Overtaken by tree-based models beyond $\sim$8 shots \\

& Tabula-8B \citep{gardner2024tabula8b}
& Text serialization (L4)
& Most reported advantage traced to contamination \citep{gorla2026illusion} \\

& LLM-FE \citep{abhyankar2025llmfe}
& Language + code (L5/6)
& Not a language-native predictor; delegates to specialized search \\

& TabPFN-TS \citep{hoo2025tabpfnts}
& Numeric ICL, synthetic prior (L8)
& Same synthetic-prior mechanism as TabPFN, transferred to time series with feature engineering rather than a time-series architecture (\cref{sec:regime-ladder}) \\

\midrule

\multirow{6}{*}{Graph}
& GraphBFF \citep{becherler2026billion}
& Numeric transformer (L8)
& Not language; relies on industrial/proprietary data \\

& Talk like a Graph \citep{fatemi2024talk}
& Text serialization (L2)
& Performance varies substantially with encoding choice (4.8--61.8 points depending on task) \\

& GraphToken \citep{perozzi2024let}
& Learned soft-prompt encoding (L7)
& Same group as Talk-like-a-Graph, opposite regime; up to 73-point gain over text encodings \\

& GraphGPT \citep{tang2024graphgpt}
& Learned projector (L7)
& Structural information is carried by a non-linguistic encoder \\

& SimKGC \citep{wang2022simkgc}
& Text serialization (L2/4)
& Gains rely on textual side information rather than graph topology \\

& G2T-FM \citep{eremeev2025turning}
& Numeric ICL + structural features, synthetic prior (L8)
& Same TabPFN-family mechanism, given graph-neighborhood features; competes with well-tuned GNNs without a graph-specific architecture \\

\midrule

\multirow{4}{*}{Time series}
& Chronos \citep{ansari2024chronos}
& Numeric tokenization (L8)
& Not language; the ``language'' framing is metaphorical \\

& TiRex \citep{nxai2025tirex}
& Numeric, recurrent (L8)
& Demonstrates the value of specialized architecture rather than LLM lineage \\

& Time-LLM \citep{jin2024timellm}
& Reprogrammed embeddings (L7)
& Ablated version outperforms the full model \citep{tan2024are} \\

& Chronicle \citep{quinlan2026chronicle}
& Jointly pretrained (not L1--8)
& Contains language by construction, so it is neither language-mediated (L1--6) nor language-free (L8); the ladder has no clean slot for jointly pretrained multimodal models. Small scale (\cref{sec:regime-ladder}) \\

\midrule

\multirow{4}{*}{Vision}
& CNN/ViT specialists
& Pixel/patch input (L8)
& Domain-specific; lacks a general language interface \\

& GPT-4V / frontier VLMs
& Learned visual tokens (L7)
& Weak performance on several low-level visual reasoning tasks \citep{fu2024blink,rahmanzadehgervi2024vision} \\

& SpatialVLM
& Learned tokens + specialist-CV labels (L7)
& Depends on specialized vision models for supervision \\

& Text-only spatial reasoning \citep{wu2024is}
& Structured serialization (L2/3)
& Outperforms image-input VLMs on synthetic layouts, but is limited to structure that can be discretized into symbolic form \\

\bottomrule
\end{tabular}
\end{lrbox}
\adjustbox{max width=\textwidth,max totalheight=0.82\textheight,center}{\usebox{\wtabbox}}

\end{table}

\subsection{Benchmarks}
\label{app:benchmarks-table}

\Cref{tab:benchmarks} lists the fifteen benchmarks this review relies on most, selected to illustrate what each protocol can and cannot distinguish. The analysis of what the table shows -- that almost every entry evaluates predictive performance and almost none evaluates representation, computation, or replacement, and that five entries report no non-linguistic specialized baseline at all -- is given in \cref{sec:benchmarks} and is not repeated here.

\begin{table}[!htbp]
\centering
\scriptsize
\setlength{\tabcolsep}{3.5pt}
\renewcommand{\arraystretch}{0.88}

\caption{Benchmarks referenced across the review, selected to illustrate what each protocol can and cannot distinguish (\cref{sec:benchmarks}). \textbf{Baseline} reports whether the benchmark's standard protocol includes a non-linguistic specialized-architecture baseline (rather than only a human baseline, other language-mediated methods, or no reported baseline); ``Unclear'' marks a case not resolved by the sources reviewed in full text.}

\label{tab:benchmarks}

\begin{lrbox}{\wtabbox}
\begin{tabular}{@{}
    >{\raggedright\arraybackslash}p{0.14\textwidth}
    >{\raggedright\arraybackslash}p{0.14\textwidth}
    >{\raggedright\arraybackslash}p{0.15\textwidth}
    >{\centering\arraybackslash}p{0.08\textwidth}
    >{\raggedright\arraybackslash}p{\dimexpr 0.49\textwidth-28pt\relax}
@{}}
\toprule
\textbf{Benchmark} &
\textbf{Modality} &
\textbf{Size / scope} &
\textbf{Baseline} &
\textbf{Known limitation} \\
\midrule

TabArena / TALENT &
Tabular &
200+ real datasets &
Yes &
Skews toward small--medium IID tables where tabular foundation models already win \citep{purucker2026beyond} \\

BeyondArena &
Tabular &
142 datasets, IID + temporal + grouped shift &
Yes &
Purpose-built to expose the IID-only bias of prior benchmarks \\

GIFT-Eval &
Time series &
23 datasets, 144K series, 7 domains &
Yes &
Includes a dedicated non-leaking pretraining split, itself a response to contamination in earlier TSFM benchmarks \\

NLGraph / GraphQA &
Graph &
Synthetic reasoning tasks (connectivity, shortest path, etc.) &
\textbf{No} &
No GNN baseline in most reported comparisons; complexity confound is not isolated from encoding choice \\

GraphInfer-Bench &
Graph &
42{,}000 samples, 6 real graphs &
Yes &
Recent (2026); limited independent replication \\

BLINK &
Vision &
14 classic CV tasks reframed as VQA &
\textbf{No} &
Reports a human baseline (95.7\%), not a specialized-CV-model baseline; multiple-choice format may bottleneck scores independently of visual competence \\

MMVP &
Vision &
CLIP-blind image pairs &
\textbf{No} &
Diagnostic comparison of VLMs with each other and with CLIP similarity judgments, not a specialized-CV baseline; generalization to non-CLIP encoders is untested \\

MLE-Bench (Lite) &
Tabular / AutoML agents &
Kaggle competition subset &
\textbf{No} &
Scored against human Kaggle medal thresholds, not a non-agentic specialized baseline; the ``Lite'' subset is easier than the full benchmark, so medal rates may not transfer \\

V4FinBench \citep{kostrzewa2026v4finbench} &
Tabular (financial) &
Corporate/financial tabular tasks &
Yes &
Domain-specific; generalization to other tabular domains is not established \\

TML-Bench \citep{pinchuk2026tmlbench} &
Tabular / data-science agents &
Data-science agents on tabular ML tasks &
Unclear &
Agentic, Level-6 orchestration setting; conflates agent tool-use skill with underlying model capability \\

WN18RR / FB15k-237 &
Knowledge graphs &
Standard KGC benchmarks &
Yes &
Rich textual entity descriptions are available to text-based methods but not pure embedding baselines; this confounds the SimKGC comparison (\cref{sec:graph-findings}) \\

CRUXEval &
Code &
800 short Python functions &
\textbf{No} &
Execution-reasoning gap may not transfer to longer, real-world programs; no non-linguistic baseline is reported in the reviewed literature (\cref{sec:adjacent-findings}) \\

MoleculeNet \citep{wu2018moleculenet} &
Chemistry &
700K+ compounds, 17 datasets across 4 categories &
Yes &
Predates language-mediated methods; \citeauthor{ross2022large}'s ten-benchmark subset is not independently re-audited for contamination here \\

ProteinGym \citep{notin2023proteingym} &
Protein &
250+ deep mutational scanning assays &
Yes &
Zero-shot fitness prediction only for most assays; \citet{su2024saprot}'s ten-task suite extends beyond this benchmark alone \\

ModelNet40 \citep{wu2015shapenets} &
Point clouds &
12{,}311 synthetic CAD models, 40 categories &
Yes &
Synthetic and clean; \citet{zhang2025point}'s binary-spatial-relation failures are documented on separate, more realistic benchmarks \\

\bottomrule
\end{tabular}
\end{lrbox}
\adjustbox{max width=\textwidth,max totalheight=0.70\textheight,center}{\usebox{\wtabbox}}

\end{table}

\subsection{Confound-control audit}
\label{app:confounds}

\Cref{tab:confound-audit} codes the three confounds of \cref{sec:comparison-scientific} -- data regime, compute budget, and information availability -- for the twelve comparisons this review relies on most.

Compute budget is the factor least often reported: only four of the twelve studies control or report it. Three of those four attach an explicit quantity to the comparison rather than only holding the model fixed, and in each case the measurement was decisive for how the comparison should be read: \citeauthor{tan2024are}'s up-to-three-orders-of-magnitude training- and inference-time measurement, \citeauthor{brehmer2024does}'s compute-matched scaling curves, and TabPFN's $230\times$ end-to-end speedup over tuned AutoML systems.

Information availability is the factor most often \emph{violated} rather than simply unreported: four of the twelve comparisons -- Tabula-8B's original claim, SimKGC, MoLFormer, and TabPFN's own comparison to its tuned baselines -- rest on an information or pretraining-data asymmetry that the source study either does not control or does not fully disclose. The comparisons this review treats as its strongest evidence are exactly the small minority that control or report all three factors at once. \Cref{app:rob} extends this coding into the six-domain SRB-6 instrument applied to all 43 primary studies.

\begin{table}[t]
\centering
\scriptsize
\setlength{\tabcolsep}{3.5pt}
\renewcommand{\arraystretch}{0.88}

\caption{Confound-control audit for the twelve most load-bearing comparisons discussed in \cref{sec:methods,sec:evaluation}, extending \cref{sec:comparison-scientific}. ``Yes'' means the source study (or a subsequent audit) reports evidence that the factor was matched or explicitly measured; ``No'' means the source study or a subsequent audit found it was \emph{not} matched; ``Not reported'' means neither the original study nor this review's search found sufficient information to code the cell either way. Absence of a report is not evidence that the factor was mishandled, only that it was not disclosed.}

\label{tab:confound-audit}

\begin{lrbox}{\wtabbox}
\begin{tabular}{@{}
    >{\raggedright\arraybackslash}p{0.27\textwidth}
    >{\centering\arraybackslash}p{0.105\textwidth}
    >{\centering\arraybackslash}p{0.12\textwidth}
    >{\centering\arraybackslash}p{0.105\textwidth}
    >{\raggedright\arraybackslash}p{\dimexpr 0.4\textwidth-28pt\relax}
@{}}
\toprule
\textbf{Comparison} &
\textbf{Data regime} &
\textbf{Compute budget} &
\textbf{Information} &
\textbf{Note} \\
\midrule

Tan et al.\ ablation: Time-LLM/GPT4TS/CALF vs.\ language-free variants \citep{tan2024are} &
Yes &
Reported (up to $10^3\times$ time) &
Yes &
This review's gold-standard controlled comparison; all three factors matched or directly measured. \\

Fatemi et al.\ encoding-sensitivity ablation \citep{fatemi2024talk} &
Yes &
Yes (same model) &
Yes &
Graph, task, and model held fixed; only textual encoding varied. \\

Brehmer et al.\ equivariant vs.\ non-equivariant transformer \citep{brehmer2024does} &
Yes &
Yes (matched compute curve) &
Yes &
Fully controlled, but neither arm is language-mediated (\cref{sec:scaling}). \\

\citeauthor{purucker2026beyond}'s BeyondArena: FMs vs.\ tree/deep-learning baselines \citep{purucker2026beyond} &
Yes (purpose-built for this) &
Not reported &
Yes &
Explicitly constructed to expose IID-only bias in prior benchmarks (\cref{tab:benchmarks}). \\

TabLLM vs.\ gradient-boosted trees, few-shot \citep{hegselmann2023tabllm,huertas2024gradient} &
Yes (few-shot is the studied variable) &
Not reported &
Yes &
Data regime is the axis under test; compute cost of the LLM call itself is not reported. \\

Tabula-8B vs.\ XGBoost/TabPFN, original claim \citep{gardner2024tabula8b,gorla2026illusion} &
Partial (few-shot claimed) &
Not reported &
\textbf{No} &
Train--test contamination was subsequently identified; instruction tuning alone reaches 92.2\% of Tabula-8B's own mean accuracy, not 92.2\% of its reported margin over the baselines. \\

SimKGC vs.\ TransE/ComplEx/RotatE \citep{wang2022simkgc} &
Not reported &
Not reported &
\textbf{No} &
Textual entity/relation descriptions were available to SimKGC but not to the structural baselines (\cref{sec:graph-findings}). \\

MoLFormer vs.\ supervised/self-supervised GNN baselines \citep{ross2022large} &
\textbf{No} &
Not reported &
\textbf{No} &
1.1-billion-molecule domain-specific pretraining was not matched by the graph baselines. \\

TabPFN vs.\ tuned AutoML systems, end-to-end cost \citep{hollmann2023tabpfn} &
N/A &
Yes ($230\times$ end-to-end speedup measured on an 18-dataset OpenML-CC18 subset) &
\textbf{No} &
Synthetic-prior pretraining is itself an information/compute investment unavailable to the baselines at deployment; the speedup is dominated by tuning cost. \\

SpatialEval: text-only input vs.\ image input, same task \citep{wu2024is} &
Yes (identical task, reformatted) &
Not reported &
Yes &
Same underlying spatial content is presented through either channel; synthetic layouts only. \\

GraphInfer-Bench: LLMs vs.\ GNNs \citep{peng2026graphinfer} &
Not reported &
Not reported &
Not reported &
Recent (2026) benchmark with limited independent replication (\cref{tab:benchmarks}). \\

Si et al.\ / Allen et al.\ feature-bias interventions \citep{si2023measuring,allen2025incontext} &
Yes (matched demonstrations) &
Not reported &
Yes &
Feature- and concept-level bias, not structural-invariance bias (\cref{sec:open-problems}, RQ2). \\

\bottomrule
\end{tabular}
\end{lrbox}
\adjustbox{max width=\textwidth,max totalheight=0.68\textheight,center}{\usebox{\wtabbox}}

\end{table}

\subsection{Extended graph evidence}
\label{app:graph-detail}

\Cref{sec:graph-findings} reports that recent comparative evaluations agree in placing language's useful contribution at the feature level rather than at the structural computation. Five are collected here. GraphInfer-Bench finds LLM-based approaches weaker than conventional GNNs on graph comparison \citep{peng2026graphinfer}. A large agentic benchmark shows LLM agents with extensive tool access, a Level 6 configuration, still struggling with sophisticated graph analysis \citep{tan2026gabench}. Lightweight GNNs outperform LLM-based reasoning for attributing failures in multi-agent systems \citep{li2026beyond}. LLM assistance improves GNN robustness mainly under labeled-data or structural deficiencies, leaving the GNN's own computation in place \citep{wang2025are}. And GraphWiz, which fine-tunes an LLM to solve graph problems through purely textual instruction-following, performs strongly on the problem types covered by its fine-tuning mixture, which leaves open whether the result reflects graph reasoning or coverage of the test distribution \citep{chen2024graphwiz}.

\subsection{Extended vision evidence}
\label{app:vision-detail}

\Cref{sec:vision-findings} reports SpatialEval's finding that text-only models outperform vision-language models given the same spatial content as an image \citep{wu2024is}. A second study isolates the same effect inside the visual channel itself, and is set out here because it is easy to misread in the opposite direction.

\citet{levental2026squares} render identical $15 \times 15$ occupancy grids two ways -- as monospaced \texttt{.}/\texttt{\#} characters, and as filled squares -- and pass both to the same visual encoder as images. Three frontier model families transcribe the character rendering at 84--91\% cell accuracy but fall to 60--73\% on the filled-square rendering of the same grid. Because neither condition is tokenized text, this is not evidence about language as a substrate at all: it isolates a discrepancy between a model's text-recognition pathway and its native visual pathway, both of which sit at Level 7. It is reported because that is a sharper localization of the vision--language bottleneck than the accuracy gap alone suggests.

Three caveats bound how much weight the result can bear. It is a single-author preprint; the models were queried through consumer interfaces rather than APIs; and the F1 gap it reports (29--39\% versus 63--84\%) is considerably starker than the accuracy gap, because accuracy is inflated by the many easily classified empty cells. Taken together with SpatialEval, the evidence supports language as a workable substrate for already-discretized structure without establishing how far that advantage extends.

\subsection{Extended adjacent-modality evidence}
\label{app:adjacent-detail}

\Cref{sec:adjacent-findings} reports the five adjacent modalities as directional, because the evidence base for each is thin. This subsection records how thin, and where the qualifications lie.

\paragraph{Chemistry: what MoLFormer's headline claim does and does not cover.} MoLFormer reports outperforming supervised and self-supervised graph neural network baselines on several downstream tasks drawn from ten molecular-property benchmark datasets, while performing competitively on two others \citep{ross2022large}. The split across task types is uneven in a way a single headline number obscures: MoLFormer leads on every reported regression benchmark, including against geometry-aware graph baselines, but on classification it takes first place on roughly half the datasets and second place on the rest, and two of the strongest baselines it is compared against are not graph networks at all. The claim supported is therefore that a sequence representation can functionally substitute for a graph-specific architecture at sufficient scale, on the specific tasks and benchmark subset where the advantage holds -- not uniform superiority across all ten benchmarks.

MoLFormer's own authors report an attention-based analysis suggesting the model recovers interatomic spatial relationships from SMILES alone. That is a structure-preservation probe in the sense of \cref{sec:distinction}, so evidence on this point is not simply absent. It is, however, an analysis by the model's own authors rather than an independently audited invariance test in the sense of \cref{sec:metrics}, and it is treated throughout as contested positive evidence rather than a resolved structural-replacement result. The countervailing evidence within the same modality is that smaller SMILES-only models exhibit structural blindness to rings and branching, with performance recovering only after graph-based information is reintroduced \citep{bougiatiotis2026improving}, and that general-purpose LLMs remain insufficient for many scientific chemistry tasks, with recent systems increasingly orchestrating specialized tools they do not replace \citep{han2025from}.

\paragraph{Coverage counts.} Counting the distinct works cited within each modality's own discussion in \cref{sec:methods}, the four core modalities draw on 21 references for graphs, 16 for tabular data, 15 for time series, and 11 for vision. The five adjacent modalities draw on three apiece for chemistry and knowledge-graph completion, two each for point clouds and protein structure, and one for code. The adjacent-modality conclusions therefore rest on a handful of studies each. Code in particular rests on a single benchmark, which is why \cref{app:boundary} records its regime boundary as untested rather than confirmed, and why \cref{sec:benchmarks} records the corresponding absence of a non-linguistic baseline.

\section{Extended Theoretical Results}
\label{app:theory}

This appendix expands four bodies of theoretical and empirical work that \cref{sec:evaluation} summarizes in a sentence apiece: the dispute over what mechanism in-context learning actually implements (\cref{app:icl-theory}), the prompt-sensitivity literature (\cref{app:prompt-sensitivity}), the pretraining distribution as a locus of structural bias (\cref{app:pretraining}), and the scaling-law evidence for structured-data foundation models (\cref{app:scaling}). The first three bear on the \emph{compute} and \emph{learn} questions of \cref{sec:distinction} specifically; the fourth explains why the fitted exponents in this literature cannot responsibly be pooled.

\subsection{What mechanism does in-context learning implement?}
\label{app:icl-theory}

\Cref{sec:theory} reports two positive formal results: transformers can be minimax-optimal for function classes such as smooth nonparametric regression and Bayesian posterior mixtures when the pretraining distribution is matched to the target tasks \citep{kim2024transformers}, and the posterior-variance component of in-context risk decays exponentially with the number of demonstrations under stated assumptions \citep{wakayama2025icl}. Together these provide a theoretical account for the positive results this review documents in low-data settings such as tabular prediction (\cref{sec:tabular-findings}).

That positive picture is contested, and a dedicated search for counterevidence located a direct challenge. A body of theoretical work argues that in-context regression succeeds because transformers implement a known algorithm -- ordinary least squares, or gradient descent -- during the forward pass. The argument has been framed as implicit meta-optimization \citep{dai2023why}, implicit Bayesian inference \citep{xie2022explanation}, provable in-context algorithm selection \citep{bai2023transformers}, and a direct demonstration that trained transformers learn linear models in-context \citep{zhang2024trained}. A related but distinct account frames the same phenomenon as implicit structure induction over the demonstrations, with no algorithm executed at all \citep{hahn2023theory}.

\citet{hill2025transformers} report empirical evidence against this specific mechanistic story: transformers trained for in-context least-squares regression fail to generalize once the prompt distribution shifts, a pattern inconsistent with genuinely implementing OLS, and their behavior instead correlates with spectral signatures of the training distribution, consistent with memorization in place of algorithm execution. Parallel evidence from mathematical reasoning points the same direction. Language models solving arithmetic problems appear to rely on a patchwork of memorized heuristics in place of a general algorithm \citep{nikankin2025arithmetic}, and performance on grade-school math word problems degrades under superficial symbolic perturbations that leave the underlying problem structure unchanged \citep{mirzadeh2024gsmsymbolic}, which is difficult to reconcile with genuine algorithmic execution.

This work does not contradict \citeauthor{kim2024transformers}'s minimax-optimality result directly: the two examine different function classes and different notions of what implementing an algorithm requires. It is, however, a located counterexample to the stronger informal claim, made elsewhere in this literature, that strong in-context regression performance is evidence the model has learned a generalizable procedure rather than a distribution-specific shortcut. The theoretical literature on why in-context learning succeeds is not settled even for the function classes where it is reported to succeed.

\subsection{Sensitivity to superficial prompt properties}
\label{app:prompt-sensitivity}

Sensitivity to superficial properties of the prompt, in place of its structural content, recurs across the in-context-learning literature. Few-shot accuracy varies substantially with the order in which otherwise-identical demonstrations are presented \citep{lu2022fantastically}, and models systematically underuse information placed in the middle of a long context relative to its beginning or end \citep{liu2024lost}. Performance on rare or long-tail patterns is better explained by their frequency in the pretraining distribution than by their intrinsic difficulty \citep{mccoy2024embers}. And the effective context length a model can use productively depends on architecture-specific sequential correlations rather than being a fixed property of the context window \citep{letey2026sequential}.

Two results are especially direct on this point. \citet{webson2022do} show experimentally that language models can learn effectively from prompts whose stated content is actively misleading, indicating that a prompt's semantic content is not reliably coupled to the computation the model performs. \citet{yoon2025rotor} report a directly relevant engineering finding: prompting and formatting alone were insufficient, in their experiments, to reliably induce order-invariant behavior, and achieving it required modifying the positional mechanism. Neither is a theorem; both are consistent with, and offer empirical support for, the distinction in \cref{sec:distinction} between describing a structural concept and computationally implementing it. \citeauthor{yoon2025rotor}'s result specifically locates where architecturally encoded bias would need to be restored -- the positional mechanism -- once description and prompting are shown insufficient on their own. It is the evidential basis for the answer given to the second central question in \cref{sec:direct-answers}, and for the second premise of the mechanistic argument in \cref{sec:mechanistic-convergence}.

\subsection{The pretraining distribution as a locus of structural bias}
\label{app:pretraining}

The pretraining distribution is a further candidate locus for structural bias, and the evidence is again a mix of formal and empirical results applying to narrower settings than in-context learning in general.

\citet{raventos2023pretraining} identify the phenomenon directly for linear regression: below a critical level of pretraining task diversity, in-context learning remains tied to a small set of pretraining tasks and behaves as if performing Bayesian model selection among them, while above it the learned procedure approaches the Bayes-optimal ridge-regression estimator. \citet{goddard2025when} extend this into a full phase diagram built from simulated regression tasks, empirically characterizing the same transition across a wider range of conditions: below a critical level, in-context learning remains tied to the pretraining task distribution; above it, the learned procedure generalizes to a broader task space. Their measured threshold -- roughly $120^\circ$ of task-angle diversity, shifting to approximately $135^\circ$ under added label noise -- is what makes RQ3 in \cref{app:design} testable against a structural-complexity axis rather than a purely qualitative claim of sharpness.

Widening the pretraining distribution is not, however, a free way to acquire structure. \citet{azizian2025how} prove that the choice of pretraining-task distribution carries a formal trade-off between robustness to distribution shift and sample efficiency: it substitutes one cost, task-specific data, for another, a less sample-efficient prior. A complementary meta-learning result derives a provable data-scaling law relating the number of pretraining tasks to generalization error through a complexity-minimization argument, giving a formal account of how quickly the benefits of task diversity accrue rather than only whether they accrue \citep{fukuchi2026provable}.

Universal-approximation results establish that transformer-based in-context learning is expressive enough, in principle, to represent a broad class of functions \citep{li2025transformers}. Expressivity is a different property from the sample and compute efficiency the results above address, and this gap -- between what transformers can represent and what they can efficiently learn -- is exactly where the current theoretical literature remains incomplete. No result surveyed here characterizes the sample complexity of learning a permutation-invariant or graph-structured function through pretraining-distribution diversity alone, which is the missing connective theory \cref{sec:precise-gap} treats as this review's central open question.

\subsection{Scaling behavior of structured-data foundation models}
\label{app:scaling}

\Cref{sec:scaling} restates one figure numerically and declines to tabulate the rest. This subsection records why. Four studies report scaling behavior for structured-data foundation models, and no two measure the same quantity, over the same range, against the same baseline.

\citet{ma2025tabdpt} fit a power-law exponent of approximately $0.4$ for tabular in-context learning, with respect to parameter count and pretraining data size considered separately. For time series, \citet{yao2025towards} report that encoder-only architectures scale more favorably than decoder-only ones, and, a distinct and important observation that changes improving in-distribution performance can reduce out-of-distribution scalability; a separate large-scale study fits power-law curves against parameter count and reports a similarly sub-linear pattern without the architecture-family comparison \citep{edwards2024scaling}. In graph learning, \citet{liu2024towards} find model depth rather than parameter count to be the dominant determinant, so the relevant scaled quantity is not even the same variable across studies. And \citet{tay2023scaling} show more broadly that the architecture with the best scaling behavior changes as scale increases, so an exponent measured over one scale range need not extrapolate beyond it.

Placing these four exponents on a single axis of ``fitted exponent'' would set incommensurable quantities side by side and invite exactly the pooled reading \cref{sec:scaling} argues against. \citeauthor{ma2025tabdpt}'s $\alpha \approx 0.4$ is therefore the only figure restated numerically in the main text, and it is restated for one modality and one model family. The substantive point is unaffected by the disaggregation: none of the four measures $\Delta(N)$, the gap to a matched-bias specialized architecture, which is the quantity the review's central question requires.

\section{Synthesis Instruments: Metrics, Failure Modes, Claim Map, and Open Problems}
\label{app:synthesis-tables}

\subsection{Evaluation metrics}
\label{app:metrics-table}

\Cref{tab:metrics} enumerates the evaluation metrics used across the literature reviewed in \cref{sec:methods}, together with the blind spot each carries. The hierarchy the table implies, output performance, generalization behavior, representation fidelity, computational behavior, and mechanistic replacement, is analyzed in \cref{sec:metrics} and is not repeated here. The single entry designed to operationalize the description-versus-computation distinction directly, the invariance-violation metric $\max_\pi \lvert f(x) - f(\pi(x))\rvert$, is used by only two papers in the entire corpus.

\begin{table}[t]
\centering
\small

\caption{Evaluation metrics used across the reviewed literature. Accuracy-based metrics dominate but cannot, by construction, distinguish ``the model computed the right structure'' from ``the model reached the right answer by another route''; the direct invariance-violation metric is the only one in this table designed specifically to make that distinction (\cref{sec:distinction}).}

\label{tab:metrics}

\begin{tabularx}{\textwidth}{@{}
    >{\raggedright\arraybackslash}p{0.20\textwidth}
    >{\raggedright\arraybackslash}X
    >{\raggedright\arraybackslash}X
@{}}
\toprule
\textbf{Metric} &
\textbf{What it measures} &
\textbf{Blind spot} \\
\midrule

Held-out accuracy / F1
& Whether predictions match labels on unseen examples
& Cannot distinguish memorization or format familiarity from genuine structural learning (\cref{sec:tabular-findings}) \\

Zero-shot / few-shot delta
& Advantage over a baseline with less adaptation
& Confounded by pretraining-corpus contamination unless a dated, held-out split is used \\

Ablation delta (component removed)
& Causal contribution of a specific model component
& Rarely reported; the single instance that matters most in this review (\citealt{tan2024are}) is an outlier in the literature, not the norm \\

Fitted scaling-law exponent
& How loss decreases with model/data scale
& Says nothing about whether the \emph{gap to a matched-bias architecture} shrinks: exponents are absolute, not relative (\cref{sec:scaling}) \\

Direct invariance violation, $\max_\pi \lvert f(x) - f(\pi(x)) \rvert$
& Whether output is literally invariant under a relevant transformation
& The only metric here that operationalizes the description-vs-computation distinction directly; used by only two papers in this review (RoToR, Set-LLM) \\

Contamination-adjusted accuracy
& Accuracy after removing detected train/test overlap
& Detection methods are themselves imperfect and likely undercount contamination \citep{golchin2024time} \\

\bottomrule
\end{tabularx}
\end{table}

\subsection{Recurring failure modes}
\label{app:failure-modes-table}

\Cref{tab:failure-modes} organizes the recurring limitations of language-mediated learning by underlying mechanism: whether the failure is plausibly resolved by scale, by a change of representation, or only by restoring specialized computation. Two of the four mechanisms, order sensitivity and the compositional ceiling, have formal rather than merely empirical grounds, and that distinction is what the mechanism-based organization is designed to expose.

\begin{table}[t!]
\centering
\small
\setlength{\tabcolsep}{5pt}

\caption{Recurring limitations of language-mediated learning, organized by underlying mechanism, not by surface symptom. Across the four core modalities, the same mechanisms recur, suggesting that these limitations arise from the representational regime rather than from any single domain.}

\label{tab:failure-modes}

\adjustbox{max width=\textwidth,max totalheight=0.86\textheight,center}{%
\begin{tabularx}{\textwidth}{@{}
    >{\raggedright\arraybackslash}p{0.18\textwidth}
    >{\raggedright\arraybackslash}p{0.27\textwidth}
    >{\raggedright\arraybackslash}p{0.25\textwidth}
    >{\raggedright\arraybackslash}X
@{}}
\toprule
\textbf{Mechanism} &
\textbf{Underlying cause} &
\textbf{Primary impact} &
\textbf{Evidence} \\
\midrule

\textbf{Order sensitivity}
& Positional encoding and causal ordering introduce dependence on the presentation order of the input.
& Set- and graph-structured inputs
& Provably non-permutation-invariant \citep{egressy2025setllm}; encoding choices can cause performance swings of 4.8--61.8 points depending on task \citep{fatemi2024talk}, with failures on isomorphic graphs \citep{thushalika2026detecting}. \\[0.8em]

\textbf{Compositional ceiling}
& A single softmax-attention layer of sub-polynomial size has a precision-independent split-VC-dimension ceiling for specific relational computations.
& Multi-hop reasoning, execution semantics, and deep feature interactions
& Theoretical limitation proven for one-layer attention on three relational tasks \citep{kozachinskiy2025strassen}; empirically reflected in accuracy collapse with reasoning depth \citep{dziri2023faith} and in failure to compose two separately learned synthetic facts, though composition succeeds when one hop is a naturally acquired fact \citep{balesni2024twohop}. \\[0.8em]

\textbf{Benchmark contamination}
& Pretraining can expose models to popular benchmarks; instruction tuning may therefore reproduce apparent structural competence without learning the underlying structure.
& Few-shot tabular and time-series benchmarks
& Two independent tabular audits report substantial contamination effects \citep{gorla2026illusion,silvestri2026when}; analogous leakage is documented for time-series benchmarks \citep{meyer2026rethinking}. \\[0.8em]

\textbf{Vision--language bottleneck}
& Information retained by the visual encoder is not necessarily expressed or exploited by the language decoder.
& Spatial and quantitative VLM reasoning
& The information bottleneck is directly localized \citep{zhang2026grid2matrix} and persists across encoder, positional-encoding, and objective variants \citep{alam2026spatial}. \\

\bottomrule
\end{tabularx}%
}
\end{table}

\subsection{Claim map and net synthesis judgment}
\label{app:claim-map-table}

\begin{figure}[t]
\centering
\includegraphics[width=0.85\textwidth]{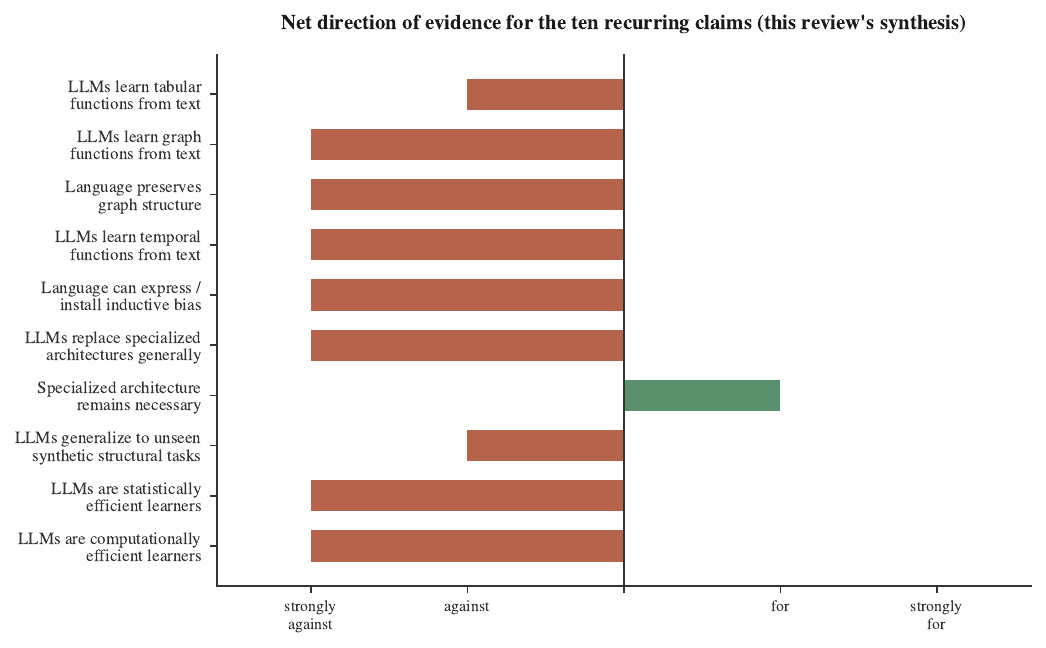}
\caption{Net direction of this review's synthesis judgment for the ten claims in \cref{tab:claim-map}. Bar direction (for/against) and magnitude (strongly/moderately) are ordinal only, deliberately restricted to two magnitude levels so that bar length cannot be read as a cardinal score derived from a citation count; the underlying judgment is qualitative and stated in full in \cref{tab:claim-map}.}
\label{fig:claimbalance}
\end{figure}

\Cref{tab:claim-map} pairs each of the ten recurring claims with the strongest supporting and counter-evidence located for it, and \cref{fig:claimbalance} renders the resulting net judgment. The three tiers into which these claims fall, and the reason nine of the ten skew against broad substitution, are set out in \cref{sec:comparative-synthesis}.

\begin{table}[t]
\centering
\small

\caption{Claim map: the ten claims in the literature, with the strongest evidence and counter evidence located on each side (\cref{sec:comparative-synthesis}). ``Status'' is this review's synthesis judgment. ``---'' marks a search that was conducted deliberately but did not locate qualifying evidence on that side, not a search that was skipped.}

\label{tab:claim-map}

\begin{tabularx}{\textwidth}{@{}
    >{\raggedright\arraybackslash}X
    >{\raggedright\arraybackslash}X
    >{\raggedright\arraybackslash}X
    >{\raggedright\arraybackslash}p{0.09\textwidth}
@{}}
\toprule
\textbf{Claim} &
\textbf{Supporting evidence} &
\textbf{Counter-evidence} &
\textbf{Status} \\
\midrule

LLMs learn tabular functions from text
& TabLLM, few-shot
& \citet{grinsztajn2022why}; \citet{gorla2026illusion}
& Narrow yes \\

LLMs learn graph functions from text
& GraphText \citep{zhao2023graphtext} (uncorroborated)
& \citet{egressy2025setllm}; \citet{thushalika2026detecting}
& Mostly no \\

Language preserves graph structure
& ---
& 4.8--61.8-point encoding-dependent swing \citep{fatemi2024talk}
& No \\

LLMs learn temporal functions from text
& LLMTime, zero-shot
& \citet{tan2024are}
& Mostly no \\

Language can express / install inductive bias
& Decomposition prompting raises SCAN success from 16.2\% to 99.7\% for the same model \citep{zhou2023leasttomost}
& \citet{yoon2025rotor}; \citet{webson2022do}
& Largely no \\

LLMs replace specialized architectures generally
& ---
& Entire method-comparison table (\cref{tab:method-comparison})
& No \\

Specialized \emph{architecture} remains necessary
& PatchTST \citep{nie2023time}, DLinear \citep{zeng2023are}, Set Transformer \citep{lee2019set}, GraphBFF \citep{becherler2026billion}
& TabPFN's synthetic-prior in-context learner transfers to time series \citep{hoo2025tabpfnts} and, with structural features, beats well-tuned GNNs \citep{eremeev2025turning} without task-specific architecture
& Narrowed \\

LLMs generalize to unseen synthetic structural tasks
& \citet{garg2022what} (simple function classes)
& \citet{dziri2023faith}; NP-complete graph tasks
& Bounded \\

LLMs are statistically efficient learners
& TabLLM extreme few-shot
& \citet{garnelo2026why}; \citet{grinsztajn2022why}
& No \\

LLMs are computationally efficient learners
& ---
& \citet{tan2024are}: up to $10^3\times$ training/inference time for no gain
& No \\

\bottomrule
\end{tabularx}
\end{table}

\subsection{Open problems}
\label{app:open-problems-table}

\Cref{tab:open-problems} restates the gap of \cref{sec:precise-gap} as five independently addressable sub-problems, each paired with the reason it remains open and a concrete experimental direction. The first row is the central gap statement; the remaining four are narrower and can be pursued separately, and each is elaborated as a research question in \cref{app:design}.

\begin{table}[t]
\centering
\small

\caption{Open problems this review identifies for future directions (\cref{sec:open-problems}). The first row is the paper's central gap statement; the remaining rows are narrower, independently addressable sub-problems.}

\label{tab:open-problems}

\begin{tabularx}{\textwidth}{@{}
    >{\raggedright\arraybackslash}X
    >{\raggedright\arraybackslash}X
    >{\raggedright\arraybackslash}X
@{}}
\toprule
\textbf{Problem} &
\textbf{Why unresolved} &
\textbf{Promising direction} \\
\midrule

No cross-regime, information-controlled measurement of the architecture-vs-language gap
& Every existing study varies modality \emph{and} representation \emph{and} real-world confounds simultaneously
& Synthetic task families with known ground truth, five matched representation arms, and a direct invariance-violation metric (\cref{sec:experimental-design}) \\

No closed-form connection between proven complexity-theoretic ceilings and an empirical structural-complexity variable
& Kozachinskiy et al.'s VC-dimension result and Goddard et al.'s phase transition are each demonstrated in isolation, on different task families
& Replicate the compositional-task impossibility result directly, extended with instruction and demonstration arms not covered by the original proof \\

Knowledge-graph completion's positive result is unexplained mechanistically
& SimKGC's win is attributed to textual side-information post hoc, not tested directly
& An ablation removing entity/relation text descriptions while preserving triples, to isolate information availability from structural inference \\

Scaling-law exponents are reported in absolute terms, never relative to a matched-bias baseline
& No paper fits a joint curve of (architecture accuracy $-$ language accuracy) against scale
& Report scaling curves as a single gap-vs-scale plot in place of two separate loss-vs-scale plots \\

Level-6 orchestration systems are evaluated on task success, not on what fraction of the computation the LLM itself performs
& Production AutoML-agent papers report end-task metrics, not a decomposition of LLM vs.\ tool contribution
& An ablation analogous to \citet{tan2024are}, applied to orchestration agents rather than forecasters \\

\bottomrule
\end{tabularx}
\end{table}

\section{Research Questions and the Minimal Experimental Design}
\label{app:design}

This appendix expands \cref{sec:rqs}. Each research question is stated with its competing hypotheses, the experiment that separates them, the observation that would falsify the hypothesis this review favors, and the confound that must be controlled for the observation to count. \Cref{app:min-design} then specifies the single protocol intended to resolve all four.

\subsection{RQ1: Does depth remove the compositional ceiling?}
\label{app:rq1}

\emph{Question.} For a fixed synthetic function class, how does the sample efficiency of in-context learning compare with that of a matched specialized architecture as structural complexity -- permutation-group size, locality radius, interaction sparsity, or compositional depth -- increases?

\citet{kozachinskiy2025strassen} prove the one-layer, sub-polynomial-size case for standard softmax attention: none of three compositional tasks (three-way matching, function composition, binary-relation composition) is solvable at that size, at arbitrary numerical precision, whereas their proposed Strassen attention solves all three at one layer and constant size. Every one of their impossibility theorems is stated for a one-layer transformer, so the proof does not establish what standard, unmodified multi-layer attention can do on the same tasks. Their own discussion of limitations does not raise the multi-layer case; the gap is noted here as an observation about the theorem's scope, not as a question the authors pose.

\begin{description}
\item[H1 (architectural ceiling).] The impossibility persists at any constant depth for standard attention of sub-polynomial size, so the barrier reflects what softmax attention computes, not the depth or width of the network.
\item[H2 (depth suffices).] Additional layers, without changing the attention mechanism itself, recover the missing compositional operations, making the one-layer result a depth limitation rather than a limitation of the mechanism.
\item[Distinguishing prediction.] As depth increases at fixed parameter count, H1 predicts no improvement on any of the three tasks; H2 predicts improvement attributable to depth before parameter count alone would explain it.
\item[Falsification of H1.] A many-layer standard-attention model closing the gap on any of the three tasks without a Strassen-style architectural change.
\item[Confound.] An apparent depth effect caused by the additional parameters introduced by increasing depth; controllable by matching parameter count across depths.
\end{description}

\subsection{RQ2: Does stating an invariance in language reduce its violation?}
\label{app:rq2}

\emph{Question.} When a structural invariance is stated explicitly in language, does the resulting output distribution become measurably closer to invariant under the direct metric of \cref{sec:metrics}?

This is narrower than it first appears, because two closely related versions have already been tested. \citet{si2023measuring} construct underspecified in-context demonstrations in which two features are equally predictive and add natural-language instructions designed to impose a bias toward one; they find such interventions can shift the learner, but that strong pretraining priors are difficult to override -- a \emph{feature-level} inductive-bias result. \citet{allen2025incontext} hold a target concept fixed and vary only the channel through which it is conveyed, image versus text, measuring how the resulting in-context inductive bias differs across vision-language models. Neither study uses a direct structural-invariance-violation metric of the kind defined in \cref{sec:metrics}, a structural-complexity sweep, or a matched specialized baseline; both concern whether language can bias a learner toward one feature or concept among several, not whether a stated \emph{invariance} -- permutation, relabeling, or relational structure -- measurably reduces the model's own violation of it. RQ2 is therefore re-scoped from an unqualified absence to a specific remaining gap: \textbf{measured for feature-level and concept-level bias, unmeasured for structural invariance under a direct violation metric.}

\begin{description}
\item[H1 (bias is inert for structural invariance specifically).] The residual gap under the direct metric remains flat or increases with dimensionality, consistent with \citeauthor{yoon2025rotor}'s prompting-only attempt before positional-encoding surgery was required (\cref{app:prompt-sensitivity}).
\item[H2 (bias is dilute but real).] Instruction narrows the gap gradually, at a rate too small to have been detected in \citeauthor{yoon2025rotor}'s particular tasks but detectable in aggregate.
\item[Critical experiment.] Measure the invariance metric directly across a dimensionality sweep, without inferring invariance from downstream task accuracy or from the feature-selection paradigms of Si et al.\ and Allen et al., neither of which can distinguish H1 from H2 at the required resolution.
\item[Falsification of H1.] A measurable narrowing that remains stable across dimensionality.
\end{description}

\subsection{RQ3: Do demonstrations close more of the gap than instruction alone?}
\label{app:rq3}

\emph{Question.} Do demonstrations close more of the gap than explicit instruction alone, and can their effect be distinguished from memorization?

\begin{description}
\item[H1 (in-support generalization).] \citeauthor{wakayama2025icl}'s exponential-decay result predicts that demonstrations provide rapidly diminishing risk when the target structure lies within the support of the pretraining task mixture.
\item[H2 (out-of-support failure).] \citeauthor{goddard2025when}'s phase transition instead predicts a qualitative break once the target structure leaves that support, rather than a smooth continuation of H1's decay. In their linear-function setting, a transformer pretrained on tasks spanning less than approximately $120^\circ$ of task-angle diversity fails on unseen tasks outside that span, whereas diversity beyond roughly $120^\circ$ -- shifting to approximately $135^\circ$ under added label noise -- yields a solution that generalizes across the full test range. This is a measured threshold, not merely a qualitative claim of sharpness (\cref{app:pretraining}).
\item[Critical experiment.] Vary the target task's distance from the pretraining mixture directly, in a setting with a matched structural-complexity axis rather than \citeauthor{goddard2025when}'s task-angle axis, and test whether the empirical curve follows H1's smooth decay or exhibits a comparable transition.
\item[Confound.] Apparent in-support generalization that is actually retrieval of a memorized near-duplicate; controllable through the contamination canary in \cref{app:min-design}.
\end{description}

\subsection{RQ4: What predicts the gap in advance?}
\label{app:rq4}

\emph{Question.} Which measurable task properties -- symmetry-group order, minimum description length, or discretizability -- predict in advance whether the gap will be small or catastrophic?

The discrete-versus-continuous split reported in \cref{sec:vision-findings}, where text-symbol grids outperform pixel grids on identical tasks while both fail on continuous geometry, provides the strongest existing empirical indication that discretizability may be an important moderator. \citeauthor{tabaghi2024universal}'s $2DN$ parameter-cost formula provides a natural, citable complexity variable against which the hypothesis can be tested.

H1. Discretizability alone predicts the gap's magnitude.
H2. Discretizability is one of several independent moderators, including symmetry-group order and minimum description length, that cannot be reduced to a single variable.
Critical experiment. Hold discretizability fixed while varying symmetry-group order and MDL independently, and test whether the gap changes.

\subsection{Minimal experimental design}
\label{app:min-design}
\label{sec:experimental-design}

The design follows directly from the gap defined in \cref{sec:precise-gap} and is intended to resolve RQ1--RQ4 through a common protocol.

\paragraph{Task families.} A small set of task families with known generating functions: permutation invariance; all three of \citeauthor{kozachinskiy2025strassen}'s compositional impossibility tasks (three-way matching, function composition, and binary-relation composition); locality; sparse tabular interaction; and graph neighborhood aggregation. The three compositional tasks are tested separately, because the one-layer proof treats them as distinct theorems and their multi-layer behavior is unknown for all three.

\paragraph{Representation arms.} For each family, five arms are compared:
\begin{enumerate}
\item a specialized architecture matched to the target bias;
\item na\"ive text serialization with no structural hint;
\item serialization plus an explicit statement of the relevant invariance;
\item serialization plus demonstrations only, with no explicit statement, isolating whether demonstrations alone can convey the bias;
\item a lossless canonical serialization, giving the best-case upper bound attainable if representation loss were the sole source of any observed gap.
\end{enumerate}

Within the taxonomy of \cref{sec:regime-ladder}, arms 2--4 occupy Levels 1--4 and arm 1 occupies Level 8. The design deliberately excludes Levels 5--7, because the question is whether the language-mediated computation itself acquires the relevant bias, not whether an orchestrated tool or separate encoder can supply it -- a question \cref{sec:autoML} and \cref{tab:replacement}'s orchestration row already answer affirmatively for a different reason.

\paragraph{Contamination control.} Fresh generating functions are sampled on every run; randomized-symbol feature names are contrasted with real-sounding names to isolate world-knowledge leakage; and one structural regime, never referenced during development, is held out as a contamination canary.

\paragraph{Measurement.} The primary dependent variable is the direct invariance-violation metric defined in \cref{sec:metrics}, not accuracy. Compute cost per prediction is reported alongside it, so that any apparent advantage remains interpretable against \citeauthor{tan2024are}'s demonstration that accuracy parity can coexist with a compute disadvantage of up to three orders of magnitude. Together these measurements populate the dimensions of \cref{tab:replacement} that end-task accuracy alone cannot: functional replacement (arm comparison at fixed accuracy), computational replacement (cost per prediction), and structural replacement (the direct invariance metric) are each measured separately.

\subsection{Concrete near-term directions}
\label{sec:future-directions}
\label{app:future-directions}

Four narrower and independently valuable directions follow directly from gaps identified in \cref{sec:methods,sec:evaluation}. None requires the full protocol of \cref{app:min-design}, and each is feasible on modest compute.

\begin{enumerate}
\item \textbf{Report structured-data scaling laws as $\Delta(N)$ curves} relative to a matched-bias baseline, in place of isolated loss-versus-scale curves. \Cref{sec:scaling} shows this connective analysis is absent from every scaling-law paper located, even though it requires no new training run and can be obtained by re-analyzing quantities most of those papers already collected.

\item \textbf{Test whether depth alone closes the compositional gap} on \citeauthor{kozachinskiy2025strassen}'s three tasks (three-way matching, function composition, binary-relation composition), keeping the attention mechanism otherwise standard and treating the tasks separately, because their proof treats them as three distinct theorems each proved only for one layer. This is feasible on the small synthetic tasks the proof already specifies, and can be run well before the harder theoretical question of a multi-layer impossibility result is resolved.

\item \textbf{Ablate SimKGC's textual side-information}, analogously to \citeauthor{tan2024are}'s time-series ablation, by removing entity and relation descriptions while holding the underlying triples fixed. This would determine how much of SimKGC's advantage over embedding-based methods arises from information availability rather than from genuine exploitation of structure, a distinction currently inferred rather than tested (\cref{sec:graph-findings}). A search for a verified prior instance of this exact test located none meeting the evidentiary standard applied elsewhere here, and both the codebase and the benchmarks are public and small.

\item \textbf{Apply a component ablation to Level-6 orchestration agents} (\cref{sec:autoML}), decomposing end-task success into the contribution attributable to the language model's search and planning versus that of the specialized tool being orchestrated. Current evaluations report primarily end-task metrics and therefore cannot distinguish genuine orchestration capability from cases in which the tool performs the substantive computation largely independently of the orchestrator's quality.
\end{enumerate}

\clearpage




\end{appendices}


\bibliographystyle{plainnat}
\bibliography{references}

\end{document}